\pdfoutput=1
\documentclass[11pt]{article}

\usepackage{EACL2023}
\usepackage{times}
\usepackage{latexsym}

\usepackage[T1]{fontenc}
\usepackage[utf8]{inputenc}

\usepackage{microtype}

\usepackage{makecell}
\usepackage{adjustbox}
\usepackage{graphicx}
\usepackage{subfig}
\usepackage{hyperref}
\usepackage{url}
\usepackage{float}
\usepackage{placeins}
\usepackage{booktabs}
\title{Does Transliteration Help Multilingual Language Modeling?}
\author{Ibraheem Muhammad Moosa \\
  Pennsylvania State University \\
  \texttt{ibraheem.moosa@psu.edu} \\\AND
   {\bf Mahmud Elahi Akhter} \and {\bf Ashfia Binte Habib}\\
  Dept. of Electrical and Computer Engineering, \\
  North South University, Dhaka, Bangladesh\\
  \texttt{\{mahmud.akhter01,ashfia.habib\}@northsouth.edu} \\\\}

\begin{document}
\maketitle
\begin{abstract}
Script diversity presents a challenge to Multilingual Language Models (MLLM) by reducing lexical overlap among closely related languages. Therefore, transliterating closely related languages that use different writing scripts to a common script may improve the downstream task performance of MLLMs. 
We empirically measure the effect of transliteration on MLLMs in this context. 
We specifically focus on the Indic languages, which have the highest script diversity in the world, and we evaluate our models on the IndicGLUE benchmark. We perform the Mann-Whitney U test to rigorously verify whether the effect of transliteration is significant or not. We find that transliteration benefits the low-resource languages without negatively affecting the comparatively high-resource languages. We also measure the cross-lingual representation similarity of the models using centered kernel alignment on parallel sentences from the FLORES-101 dataset. We find that for parallel sentences across different languages, the transliteration-based model learns sentence representations that are more similar.
\end{abstract}

\section{Introduction}
\label{introduction}
In the last few years, we have seen impressive advances in many NLP tasks. These advances have been primarily led by the availability of large representative corpora and improvement in the architecture of large language models. While improving model architectures, training methods, regularization techniques, etc., can help advance the state of NLP in general, the unavailability of large, diverse corpora is the bottleneck for most languages \citep{joshi-etal-2020-state}. Thus to inclusively advance the state of NLP across languages, it is crucial to develop techniques for training MLLMs that can extract the most out of existing multilingual corpora. Here, we focus on the issue of diverse writing scripts used by closely related languages that may prevent MLLMs from learning good cross-lingual representations. Previous papers \citep{Pfeiffer2021UNKsEA} have noted that low-resource languages that use unique scripts tend to have very few tokens representing them at the tokenizer. As a result, these languages tend to have more \textit{UNK}nown tokens, and the words in these languages tend to be more split up by sub-word tokenizers. Often we can easily transliterate from one script to another using rule-based systems. For example, there are established standards that can be used to transliterate Greek (ISO 843), Cyrillic (ISO 9), Indic scripts (ISO 15919), and Thai (ISO 11940) to the Latin script. 

In this paper, we focus on the Indic languages, which have the highest script diversity in the world. Many South Asian and Southeast Asian languages are intimately connected linguistically, historically, phonologically \citep{littell-etal-2017-uriel} and phylogenetically. However, due to different scripts, it is difficult for MLLMs to fully exploit this shared information. Among the Indic languages we considered in this study we encounter eleven different scripts. These are shown in Table~\ref{table-pr-dt}. Nevertheless, these scripts have shared ancestry from the ancient Brahmic script \citep{Hockett1997, Coningham1996} and have similar structures that we can easily use to transliterate them to a common script. %The ISO-15919 standard has been designed to perform this transliteration to the Latin script.
Also, many of these languages heavily borrow from Sanskrit, and due to its influence, many words are shared among these languages. Therefore, due to their relatedness and highly diverse script barrier, the Indic languages presents a unique opportunity to analyze the effects of transliteration on MLLMs. 

We empirically measure the effect of transliteration on the downstream performance of MLLMs. We pretrain ALBERT (Base, 11M Parameters) \citep{Lan2020ALBERTAL} and RemBERT (Base, 192M Parameters) \citep{Chung2021RethinkingEC} models from scratch on Indic languages. We pretrain two variants of each model, one with the original writing scripts and the other after transliterating to a common writing script. Henceforth, we will refer to the transliterated script model as uni-script model and the other as a multi-script model. We evaluate the models on downstream tasks from the IndicGLUE benchmark dataset \citep{kakwani-etal-2020-indicnlpsuite}. In order to rigorously compare the two models, we finetune using nine random seeds on all downstream tasks. Then we perform the Mann-Whitney U test (MWU) between the uni-script and multi-script models. Using the MWU test, we conclude that transliteration significantly benefits the low-resource languages without negatively affecting the comparatively high-resource languages.

We also measure the Cross-Lingual Representation Similarity (CLRS) to understand why the uni-script model performs better than the multi-script model. To measure the CLRS, we use the centered kernel alignment (CKA) \citep{Kornblith2019SimilarityON} similarity score. We measure the CKA similarity score between the hidden representations of the models on the parallel sentences of the Indic languages from the FLORES-101 dataset \citep{Goyal2021TheFE}. We find that, compared to the multi-script models, the uni-script models achieve a higher CKA score, and it is more stable throughout the hidden layers of the models. Based on this, we conclude that the uni-script models learn better cross-lingual representation than the multi-script models. In summary, our contributions are primarily three-fold:
\begin{enumerate}
\item We find that transliteration significantly benefits the low-resource languages without negatively affecting the comparatively high-resource languages.
\item We establish this finding through rigorous experiments and show the statistical significance along with the effect size of transliteration using the Mann-Whitney U test.
\item Using CKA on the FLORES-101 dataset, we show that transliteration helps MLLMs learn better cross-lingual representation.
\end{enumerate}
Our code is available at Github\footnote{\url{https://github.com/ibraheem-moosa/XLM-Indic}}
and our model weights can be downloaded from HF Hub
\footnote{\url{https://huggingface.co/ibraheemmoosa/xlmindic-base-uniscript}}
\footnote{\url{https://huggingface.co/ibraheemmoosa/xlmindic-base-multiscript}}
\footnote{\url{https://huggingface.co/ibraheemmoosa/xlmindic-rembert-uniscript}}
\footnote{\url{https://huggingface.co/ibraheemmoosa/xlmindic-rembert-multiscript}}.
\section{Motivation and Background}
\label{motivation-and-background}
\subsection{Motivation}
\label{motivation}
In their study, \citet{joshi-etal-2020-state} showed the resource disparity between low-resource and high-resource languages, and \citet{ruder2020beyondenglish} discussed the necessity of working with low-resource languages. A large body of work suggests that language-relatedness can help MLLMs achieve better performance on low-resource languages by leveraging related high-resource languages. For instance, \citet{pires-etal-2019-multilingual} found that lexical overlap improved mBERT’s multilingual representation capability even though it learned to capture multilingual representations with zero lexical overlaps. \citet{dabre-etal-2017-empirical} showed that transfer learning in the same or linguistically similar language family gives the best performance for NMT. \citet{lauscher-etal-2020-zero} found that language relatedness is crucial for POS-tagging and dependency parsing tasks. Although, corpus size is much more important for NLI and Question Answering tasks. \citet{wu-dredze-2020-languages} showed that bilingual BERT outperformed monolingual BERT on low-resource languages when the languages were linguistically closely related. Nevertheless, mBERT outperformed bilingual BERT on low-resource languages. 
\subsection{Script Barrier in Multilingual Language Models}
\label{barrier}
One of the major challenges in leveraging transfer between high-resource and low-resource languages is overcoming the script barrier. Script barrier exists when multiple closely related languages use different scripts. \citet{anastasopoulos-neubig-2019-pushing} found that for morphological inflection, script barrier between closely related languages impedes cross-lingual learning, and language relatedness improved cross-lingual transfer. Transliteration and phoneme-based techniques have been proposed to solve this issue. For example, \citet{murikinati-etal-2020-transliteration} expanded upon \citet{anastasopoulos-neubig-2019-pushing} and showed that both transliteration and grapheme to phoneme (g2p) conversion removes script barrier and improves cross-lingual morphological inflection and \citet{Rijhwani2019ZeroshotNT} showed that pivoting low-resource languages to their closely related high-resource languages results in better zero shot entity linking capacity and used phoneme-based pivoting to overcome the script barrier. \citet{bharadwaj-etal-2016-phonologically} showed that phoneme representation outperformed orthographic representations for NER. \citet{Chaudhary2018AdaptingWE} also used phoneme representation to resolve script barriers and adapt word embeddings to low-resource languages. 
\subsection{Transliteration in Language Modeling}
\label{transliteration-model}
Different works have applied transliteration in different aspect for language models. For instance, \citet{goyal-etal-2020-efficient} and \citet{song-etal-2020-pre} both utilized transliteration and showed that language relatedness was required for improving performance on NMT. \citet{amrhein-sennrich-2020-romanization} studied how transliteration improved NMT and came to the conclusion that transliteration offered significant improvement for low-resource languages with different scripts. 

\citet{khemchandani-etal-2021-exploiting} showed on Indo-Aryan languages that language relatedness could be exploited through transliteration along with bilingual lexicon-based pseudo-translation and aligned loss to incorporate low-resource languages into pretrained mBERT. \citet{muller-etal-2021-unseen} showed that for unseen languages, the script barrier hindered transfer between low-resource and high-resource languages for MLLMs and transliteration removed this barrier. They showed that transliterating Uyghur, Buryat, Erzya, Sorani, Meadow Mari, and Mingrelian to Latin script and finetuning mBERT on the respective corpus with masked language modeling objective improved their downstream POS performance significantly. In contrast, \citet{DBLP:conf/iclr/KWMR20} and \citet{artetxe-etal-2020-cross} proposes that mBERT can learn cross-lingual representations without any lexical overlap, a shared vocabulary, or joint training. However, these works focus on zero-shot cross-lingual transfer learning only.
From the literature, it can be seen that many in the community believe transliteration to be a potential solution for script barriers. However, most of the work shows the benefits of transliteration for NMT. Nevertheless, there is no solid empirical analysis of the effects of transliteration for MLLMs apart from \citet{dhamecha-etal-2021-role,muller-etal-2021-unseen}. Hence, the motivation behind this paper is to provide a solid empirical analysis of the effect of transliteration for MLLMs with statistical analysis and determine whether or not it helps models learn better cross-lingual representation.

It should also be noted that, even though our idea seems to be similar to \citet{muller-etal-2021-unseen} and \citet{dhamecha-etal-2021-role}, there are major differences. For instance, \citet{muller-etal-2021-unseen} adapted existing pretrained model to very low-resource languages. Whereas, we focus on training the models with transliteration from scratch. We also train our models on 20 languages and evaluate on more than 50 tasks. Unlike \citet{dhamecha-etal-2021-role}, we also include Dravidian Languages in our analysis. Furthermore, we focused on the issue of script barrier while \citet{dhamecha-etal-2021-role} focused on multilingual fine-tuning. Whereas, we adopt multilingual fine-tuning on all our models. Thus the improvement we see comes only from circumventing the script barrier. Moreover, we have provided statistical testing to show the significance of transliteration instead of just showing better metrics. We also performed cross-lingual representation similarity analysis to show the benefits of transliteration. 

\subsection{Cross Lingual Similarity Learning in Language Modeling}
\label{translit-model}
Several techniques have recently been used to study the hidden representations of multilingual language models. \citet{kudugunta-etal-2019-investigating} study CLRS of NMT models using SVCCA \citep{Raghu2017SVCCASV}. \citet{Singh2019BERTIN} used PWCCA \citep{Morcos2018InsightsOR} to study the CLRS of mBERT and found that it drastically fell with depth. \citep{Wu2020EmergingCS} have used CKA to study the CLRS of bilingual BERT models. They found that similarity is highest in the first few layers and drops moderately with depth. \citet{Muller2021FirstAT} used CKA to study CLRS of mBERT before and after finetuning on downstream tasks. They found in all cases that CLRS increases steadily in the first five layers, then it decreases in the later layers. From this, they concluded that mBERT learns multilingual alignment in the early layers and preserves it throughout finetuning. \citet{Del2021EstablishingII} applied various similarity measures to understand CLRS of various multilingual masked language models. Their results also show that CLRS increases in the first half of the models, while in the later layers, this similarity steadily falls.
\section{Experiment and Results}
\subsection{Mann–Whitney U test}
\label{statistical-analysis}
We perform Mann–Whitney U test (MWU) \citep{10.1214/aoms/1177730491, Wilcoxon1945} to determine if the performance differences between the multi-script and the uni-script models are significant. In short, it tells us the effect of transliteration on model performance. 
MWU is a non-parametric hypothesis test between two groups/populations. MWU is chosen because it has weak assumptions. The only assumptions of MWU are that the samples of the two groups are independent of each other, and the samples are ordinal. Under the MWU, our null hypothesis or \textbf{h\textsubscript{0}} is that the performances of the uni-script (group 1) and the multi-script (group 2) models are similar, and the alternative hypothesis or \textbf{h\textsubscript{a}} is that the performances (groups) are different. We set our confidence interval \(\alpha\) at 0.05 and reject the \textbf{h\textsubscript{0}} for the p-values < \(\alpha\). We also report three test statistics as the p-value only gives statistical significance, which can be misleading at times \citep{pmid23997866}.

The test statistics are three different effect sizes that convey three different information. These test statistics are absolute effect size ($\delta$), common language effect size ($\rho$), and standardized effect size ($r$). The absolute effect size $\delta$ is the difference between the mean of the models' performance metric, which is given as, 
\begin{center}$\delta$ = $\mu$\textsubscript{uni-script}-$\mu$\textsubscript{multi-script}\end{center}
for any given task and language. When the \textbf{h\textsubscript{0}} is rejected for any given task, a positive $\delta$ indicates the uni-script model is better, and a negative $\delta$ indicates the multi-script model is better. The details and results of common language effect size ($\rho$), and standardized effect size ($r$) are presented in appendix~\ref{sec:rho&stdeffect}.
\subsection{Dataset}
\label{dataset}
The ALBERT models were pretrained on a subset of the OSCAR corpus containing Indo-Aryan languages.
We use the unshuffled deduplicated version of OSCAR corpus \citep{OrtizSuarezSagotRomary2019} available via Huggingface datasets library \citep{lhoest-etal-2021-datasets}. We pretrain on Panjabi, Hindi, Bengali, Oriya, Assamese, Gujarati, Marathi, Sinhala, Nepali, Sanskrit, Goan Konkani, Maithili, Bihari, and Bishnupriya portion of the OSCAR corpus.

The RemBERT models were trained on a significantly larger pretraining corpus with additional languages. We pretrained the RemBERT models on a combination of Wikipedia \citep{wikidump}, mC4 \citep{2019t5}, OSCAR2109 \citep{AbadjiOrtizSuarezRomaryetal.2021} and OSCAR corpus. These datasets are also available via the Huggingface datasets library. In addition to the languages in the ALBERT pretraining corpus, we consider English, four Dravidian languages Kannada, Telugu, Malayalam, and Tamil, and an Indo-Aryan language Dhivehi. 
\begin{table}[hbt!]
\begin{center}
\begin{adjustbox}{max width=.45\textwidth}
\small
\begin{tabular}{l c c c c}
% \label{scripts-tb}
\\ \hline 
\textbf{Lang.} & \textbf{Sub-family} & \textbf{Script} & \textbf{Size(GB)}\\
\hline 
en & Germanic & Latin & 131\\
hi & Central Indo-Aryan & Devanagari & 43\\
mr & Southern Indo-Aryan & Devanagari & 35\\
bn & Eastern Indo-Aryan & Bengali & 28\\
ta & South Dravidian & Tamil & 22\\
ml & South Dravidian & Malayalam & 10\\
te & South-Central Dravidian  & Telugu & 7\\
kn & South Dravidian & Kannada & 6\\
si & Insular Indo-Aryan & Sinhala & 5\\
ne & Northern Indo-Aryan & Devanagari & 4\\
gu & Western Indo-Aryan & Gujarati & 3.5\\
pa & Northwestern Indo-Aryan & Gurmukhi & 2\\
or & Eastern Indo-Aryan & Oriya & 0.5\\
sa & Sanskrit & Devanagari & 0.2\\
as & Eastern Indo-Aryan & Bengali & 0.1\\
dv & Insular Indo-Aryan & Thaana & 0.1\\
bpy & Eastern Indo-Aryan & Bengali & < 0.1\\
gom & Southern Indo-Aryan & Devanagari & < 0.1\\
bh & Eastern Indo-Aryan & Devanagari & < 0.1\\
mai & Eastern Indo-Aryan & Devanagari & < 0.1\\
\hline 
\end{tabular}
\end{adjustbox}
\caption{Languages in our pretraining corpus and their writing scripts and the pretraining corpus sizes used for the RemBERT model}
\label{table-pr-dt}
\end{center}
\end{table}
We evaluate the models on four downstream tasks from IndicGLUE \citep{kakwani-etal-2020-indicnlpsuite}, which are News Article Classification, WSTP, CSQA, and NER. We use the balanced Wikiann dataset from \citet{rahimi-etal-2019-massively} for NER. In addition, we evaluate the models on other publicly available datasets that are part of the IndicGLUE benchmark. These are BBC Hindi News Classification, Soham Bengali News Classification, INLTK Headlines Classification, IITP Movie, and Product Review Sentiment Analysis \citep{cicling/Akhtar16}, MIDAS Discourse Mode Classification \citep{Dhanwal2020AnAD} and ACTSA Sentiment Classification \citep{mukku-mamidi-2017-actsa} datasets.
\subsection{Transliteration Method}
\label{transliteration-method}
We transliterate Indic language texts to Latin script using the ISO 15919 transliteration scheme. We tested with two publicly available implementations of this scheme, Aksharamukha  \citep{Aksharamukha} and PyICU \citep{pyicu}. We found the quality of transliteration of the Aksharamukha library to be better. Thus we use this library for transliterating the inputs to the ALBERT uni-script model. However, the Aksharamukha implementation is very slow compared to the PyICU implementation. As we significantly expanded our pretraining corpus for the RemBERT model, we switched to PyICU for the RemBERT uni-script model.
\subsection{Downstream Finetuning}
\label{downstream-finetuning}
We finetune the models on each downstream task independently.
The specific hyperparameters used for each task are reported in the appendix~\ref{sec:pretraining}. On all tasks, we finetune with nine random seeds and report the average and standard deviation of the metrics. In Table~\ref{table-res} and Table~\ref{table-res-csqa}, we report the performances on IndicGLUE benchmark tasks and in Table~\ref{table-res-pb} on other publicly available datasets. Here, we discuss the results on each of the the models on each of the tasks. Furthermore, in appendix~\ref{sec:rho&stdeffect}, we show the test statistics for all the datasets. 
% table Test accuracy on various MCQA tasks
\begin{table*}[hbt!]
%\begin{center}
\begin{adjustbox}{max width=\textwidth}
% \resizebox{\textwidth}{!}{
%\small
\begin{tabular}{l c c c c c c c c c c c c}
%\hline\\
\toprule[2pt]
\textbf{Model} & \textbf{pa} & \textbf{hi} & \textbf{bn} & \textbf{or} & \textbf{as} & \textbf{gu} & \textbf{mr} & \textbf{kn} & \textbf{te} & \textbf{ml} & \textbf{ta} & \textbf{avg} \\
\midrule[2pt]
\multicolumn{13}{l}{\textbf{Wikipedia Section Title Prediction}}\\
%\multicolumn{9}{l}{\textbf{RemBERT}}\\
RemBERT\textsubscript{$MS$}   & 68.42±0.92 & 70.90±0.39 & 72.58±0.45 & 69.92±0.90 & 68.37±1.37 & 72.93±0.58 & 73.23±0.61 & 71.67±0.41 & 92.98±0.19 & 69.03±0.57 & 69.77±0.45 & 73.00\\
RemBERT\textsubscript{$US$}   & 71.01±0.22 & 72.45±0.29 & 73.65±0.21 & 75.37±0.69 & 72.50±0.91 & 76.35±0.29 & 74.58±0.72 & 74.21±0.29 & 93.66±0.09 & 69.33±0.35 & 70.63±0.22 & 74.89\\
$\delta$ & \textcolor{blue}{\textbf{2.59}} & \textcolor{blue}{\textbf{1.55}} & \textcolor{blue}{\textbf{1.07}} & \textcolor{blue}{\textbf{5.45}} & \textcolor{blue}{\textbf{4.13}} & \textcolor{blue}{\textbf{3.42}} & \textcolor{blue}{\textbf{1.34}} & \textcolor{blue}{\textbf{2.54}} & \textcolor{blue}{\textbf{0.68}} & \textcolor{orange}{\textbf{0.31}} & \textcolor{blue}{\textbf{0.86}} & \textcolor{blue}{\textbf{1.89}}\\
$p-value$ &0.0004	&0.0004	&0.0004	&0.0004	&0.0004	&0.0004	&0.0035 &0.0004 & 0.0004 & 0.2505 & 0.0006 & -\\
%\multicolumn{9}{l}{\textbf{AlBert}}\\
\midrule[1pt]
ALBERT\textsubscript{$MS$}           & 74.33±0.83 & 78.18±0.33 & 81.18±0.28 & 74.35±1.2 & 76.70±0.83 & 76.37±0.53 & 79.10±0.84 & - &- & - & - & 77.17 \\
ALBERT\textsubscript{$US$}   & 77.55±0.61 & 82.24±0.18 & 84.38±0.29 & 81.47±0.99 & 81.74±0.82 & 82.39±0.27 & 82.74±0.52 & - &- & - & - & 81.78 \\
$\delta$ & \textcolor{blue}{\textbf{3.22}} & \textcolor{blue}{\textbf{4.06}} & \textcolor{blue}{\textbf{3.20}} & \textcolor{blue}{\textbf{7.12}} & \textcolor{blue}{\textbf{5.04}} & \textcolor{blue}{\textbf{6.02}} & \textcolor{blue}{\textbf{3.64}} & - & - & - & - & \textcolor{blue}{\textbf{4.61}}\\
$p-value$ &0.0004	&0.0004	&0.0004	&0.0004	&0.0004	&0.0004	&0.0004 &- &- & - & - & - \\
\midrule[2pt]
\multicolumn{13}{l}{\textbf{News Category Classification}}\\
%\multicolumn{9}{l}{\textbf{RemBERT}}\\
RemBERT\textsubscript{$MS$}  & 95.67±0.38 & - & 97.90±0.17 & 96.59±0.18 & - & 98.22±0.58 & 99.16±0.16 & 97.23±0.10 & 99.03±0.12 & 91.25±0.43 & 97.33±0.18 & 96.93 \\
RemBERT\textsubscript{$US$}  & 96.92±0.29 & - & 97.78±0.12 & 97.55±0.14 & - & 99.02±0.14 & 99.14±0.21 & 97.10±0.12 & 99.03±0.66 & 92.08±0.40 & 97.49±0.20 & 97.34 \\
$\delta$ & \textcolor{blue}{\textbf{1.24}} & - & \textcolor{orange}{\textbf{-0.11}} & \textcolor{blue}{\textbf{0.95}} & - & \textcolor{blue}{\textbf{0.80}} & \textcolor{orange}{\textbf{-0.03}} &  \textcolor{orange}{\textbf{-0.13}} & \textcolor{orange}{\textbf{0.00}} & \textcolor{blue}{\textbf{0.83}} & \textcolor{orange}{\textbf{0.16}} & \textcolor{blue}{\textbf{0.41}}\\
$p-value$ & 0.0003	& - & 0.0981 & 0.0004 & - & 0.0040 & 0.7783	& 0.0995 & 0.7548 & 0.0014 & 0.0814 & - \\
\midrule[1pt]
%\multicolumn{9}{l}{\textbf{AlBert}}\\
ALBERT\textsubscript{$MS$}            & 96.83±0.19 & - & 98.14±0.14 & 98.09±0.16 & - & 98.80±0.43 & 99.58±0.25 & - &- & - & - & 98.30\\
ALBERT\textsubscript{$US$}   & 97.90±0.17 & - & 97.99±0.22 & 98.77±0.12 & - & 99.40±0.54 & 99.47±0.21 & - &- & - & - & 98.70\\
$\delta$ & \textcolor{blue}{\textbf{1.07}} & - & \textcolor{orange}{\textbf{-0.15}} & \textcolor{blue}{\textbf{0.68}} & - & \textcolor{blue}{\textbf{0.60}} & \textcolor{orange}{\textbf{-0.18}} & - & - & - & - & \textcolor{blue}{\textbf{0.40}}\\
$p-value$ &0.0003	&-	&0.181	&0.0004	&-	&0.03084 &0.1683 &- &- & - & - & -\\
\midrule[2pt]
\multicolumn{13}{l}{\textbf{\textbf{Named Entity Recognition (F1-Score)}}}\\
%\multicolumn{9}{l}{\textbf{RemBERT}}\\
RemBERT\textsubscript{$MS$}  & 69.47±1.72 & 90.95±0.33 & 95.51±0.18 & 87.92±1.26 & 79±0.22 & 69±0.94 & 90.72±0.17 & 72.65±1.81 & 81.82±1.81 & 89.17±0.25 & 90.07±0.33 & 83.40\\
RemBERT\textsubscript{$US$}   & 81.91±1.93 & 91.73±0.39 & 96.19±0.21 & 88.92±2.88 & 83.50±2.75 & 80.25±1.42 & 90.75±0.35 & 78.98±1.50 & 84.97±0.45 & 89.26±0.46 & 90.18±0.27 & 86.97\\
$\delta$ &\textcolor{blue}{\textbf{12.44}}	&\textcolor{blue}{\textbf{0.78}}	&\textcolor{blue}{\textbf{0.68}}	&\textcolor{orange}{\textbf{1.00}}	&\textcolor{blue}{\textbf{4.28}}	&\textcolor{blue}{\textbf{10.31}}	&\textcolor{orange}{\textbf{0.02}} &\textcolor{blue}{\textbf{6.33}} &\textcolor{blue}{\textbf{3.15}} &\textcolor{orange}{\textbf{0.01}} &\textcolor{orange}{\textbf{0.12}} & \textcolor{blue}{\textbf{3.56}} \\
$p-value$ &0.00004	&0.0005	&0.00001	&0.1615	&0.0019	&0.00004	&0.6665 &0.00004 &0.00004 & 0.7304 & 0.2973 & -\\
\midrule[1pt]
%\multicolumn{9}{l}{\textbf{ALBERT}}\\
ALBERT\textsubscript{$MS$}           & 76.69±1.5 & 91.80±0.42 & 96.39±0.19 & 84.18±1.8 & 75.45±1.8 & 69.10±2.9 & 88.72±0.40 & - &- & - & - & 83.19\\
ALBERT\textsubscript{$US$}   & 85.42±1.9 & 92.93±0.21 & 97.31±0.22 & 93.54±0.58 & 89.06±2.2 & 80.16±0.15 & 90.56±0.44 & - &- & - & - & 89.85\\
$\delta$ & \textcolor{blue}{\textbf{8.73}} & \textcolor{blue}{\textbf{1.13}} & \textcolor{blue}{\textbf{0.92}} & \textcolor{blue}{\textbf{9.36}} & \textcolor{blue}{\textbf{13.61}} & \textcolor{blue}{\textbf{11.06}} & \textcolor{blue}{\textbf{1.84}} & - & - & - & - & \textcolor{blue}{\textbf{6.66}}\\
$p-value$ & 0.0004066	& 0.0004066	& 0.0003983	& 0.0004038	& 0.000401	& 0.0004066	& 0.0004095 & - & - & - & - & -\\
\bottomrule[2pt]
\multicolumn{13}{l}{\textcolor{orange}{orange} indicates the multi-script and uni-script models are equal and \textcolor{blue}{blue} indicates the uni-script model is better}
\end{tabular}%
% }
\end{adjustbox}
\caption{Results on Classification Tasks from IndicGLUE Benchmark}
\label{table-res}
%\end{center}
\end{table*}
% table Test accuracy on various MCQA tasks

\textbf{Wikipedia Section Title Prediction:}
For both RemBERT and ALBERT models, the uni-script model performed better on all languages except Malayalam (ml). We noticed that a letter of Malayalam script is not properly transliterated by the PyICU library. This introduced some artifacts in the form of unnecessary splitting of words by the subword tokenizer.

\textbf{News Category Classification:} 
It is interesting that on this task the uni-script models performed better for Panjabi (pa) and Oriya (or) languages. It is clear from Table~\ref{table-pr-dt} that these two languages are low-resource compared to Bengali (bn) and Marathi (mr). On Bengali and Marathi we see slight performance degradation which is not statistically significant. This shows the validity of our first finding.

\textbf{Named Entity Recognition:}
On this task we see that the uni-script model performs much better for Assamese (as), Oriya(or), Panjabi (pa) and Gujarati (gu). These languages are low-resource and here again the uni-script model shines. The large performance improvement on this task can be explained by the fact that Named Entities usually have the same spelling after transliteration for Indian languages. Thus the uni-script model has better chances for learning various named-entities during pre-training.

% table Test accuracies on public datasets
\begin{table*}[hbt!]
\centering
\begin{adjustbox}{max width=\textwidth}
\small
\begin{tabular}{l c| c c c c| c c c c}
\toprule[2pt]
%\textbf{Language} & \textbf{Dataset} &\multicolumn{2}{c}{\textbf{multi-script}} &\multicolumn{2}{c}{\textbf{uni-script}} &\multicolumn{2}{c}{\textbf{Test Statistics}} \\
\textbf{Language} & \textbf{Dataset}  & RemBERT\textsubscript{$MS$}           & RemBERT\textsubscript{$US$}     &\textbf{$\delta$}           & \textbf{$p-value$}      & ALBERT\textsubscript{$MS$}          & ALBERT\textsubscript{$US$} &  \textbf{$\delta$}           & \textbf{$p-value$} \\          
\midrule[2pt]
\multicolumn{8}{l}{\textbf{Article Genre Classification}}\\
hi          & BBC News  & 76.80±0.84& 77.78±0.92 & \textcolor{blue}{\textbf{0.98}} &0.0466 & 77.28±1.51 & 79.14±0.60  & \textcolor{blue}{\textbf{1.86}} & 0.0088\\
bn          & Soham News Article Classification & 92.86±0.10 & 93.69±0.20 & \textcolor{blue}{\textbf{0.83}} &0.0004 & 93.22±0.49 & 93.89±0.48  & \textcolor{blue}{\textbf{0.67}} & 0.0090\\
gu          & INLTK Headlines  & 90.27±0.47 & 91.60±0.28 & \textcolor{blue}{\textbf{1.33}} &0.0004 & 90.41±0.69 & 90.73±0.75   &\textcolor{orange}{\textbf{0.32}} & 0.6249\\
mr          & INLTK Headlines  & 91.24±0.50 & 92.27±0.39 & \textcolor{blue}{\textbf{1.03}} & 0.0008 &  92.21±0.23 & 92.04±0.47  &\textcolor{orange}{\textbf{-0.17}} & 0.3503\\
ml          & INLTK Headlines  & 94.11±0.49 & 93.33±0.22 & \textcolor{cyan}{\textbf{-0.78}} &0.003 &- &- &- &-\\
ta          & INLTK Headlines  & 95.59±0.70 & 94.93±0.30 & \textcolor{cyan}{\textbf{-0.65}} &0.013 &- &- &- &-\\
\midrule[2pt] 
\multicolumn{8}{l}{\textbf{Sentiment Analysis}}\\
hi          & IITP Product Reviews & 72.17±1.98 & 72.85±0.63 & \textcolor{orange}{\textbf{0.68}} &0.9646 & 76.33±0.84 & 77.18±0.77    & \textcolor{blue}{\textbf{0.85}} & 0.04099 \\
hi          & IITP Movie Reviews & 58.66±1.09 & 62.65±2.74 & \textcolor{blue}{\textbf{3.99}} &0.0023 & 65.91±2.2 & 66.34±0.16  & \textcolor{orange}{\textbf{0.15}} & 0.8941  \\
te          & ACTSA & 61.18±1.38 & 60.53±0.85 & \textcolor{orange}{\textbf{-0.66}} &0.1981 &- &- &- &-  \\
\midrule[2pt]
\multicolumn{8}{l}{\textbf{Discourse Mode Classification}}\\
hi          & MIDAS Discourse & 78.07±0.83 & 79.46±0.67 & \textcolor{blue}{\textbf{1.39}} &0.0415 & 78.39±0.33 & 78.54±0.91 & \textcolor{orange}{\textbf{0.15}} & 0.7561  \\
\bottomrule[2pt]
\multicolumn{8}{l}{\textcolor{orange}{orange} indicates the multi-script and uni-script models are equal, \textcolor{cyan}{cyan} indicates multi-script is better than uni-script models and \textcolor{blue}{blue} indicates vice versa}
\end{tabular}
\end{adjustbox}
\caption{Accuracy on Public Datasets}
\label{table-res-pb}
\end{table*}
%Table~\ref{table-res-pb}.
% table Test accuracy on public datasets
\begin{table*}[hbt!]
\begin{center}
\begin{adjustbox}{max width=0.8\textwidth}
% \resizebox{0.80\textwidth}{!}{
\small
\begin{tabular}{l c c c c c c c c c c c c}
%\hline\\
\toprule[2pt]
\textbf{Model} & \textbf{pa} & \textbf{hi} & \textbf{bn} & \textbf{or} & \textbf{as} & \textbf{gu} & \textbf{mr} & \textbf{ta} & \textbf{te} & \textbf{ml} & \textbf{kn} & \textbf{avg}\\ 
\midrule[2pt]
\multicolumn{13}{l}{\textbf{Cloze-style QA (Zero Shot)}}\\
%\multicolumn{9}{l}{\textbf{Independent Baseline}}\\
%IndicBERT base (Our Evaluation)          & 29.33    & 30.76  & 28.45  & 30.38  & 29.98  & 81.50  & 29.71 & 37.16   \\
%\hline
%\multicolumn{9}{l}{\textbf{Ours}}\\
RemBERT\textsubscript{$MS$}  & 33.93 & 39.06 & 38.93 & 37.32 & 37.66 & 84.21 & 46.15 & 37.02 & 34.42 & 38.45 & 40.75 & 42.53 \\
RemBERT\textsubscript{$US$}  & 33.92 & 40.10 & 39.62 & 38.28 & 39.26 & 85.37 & 45.92 & 36.68 & 34.36 & 37.16 & 44.29 & 43.17 \\
$\delta$ & \textcolor{cyan}{\textbf{-0.01}} & \textcolor{blue}{\textbf{1.04}} & \textcolor{blue}{\textbf{0.69}} & \textcolor{blue}{\textbf{0.96}} & \textcolor{blue}{\textbf{1.6}} & \textcolor{blue}{\textbf{1.16}} & \textcolor{cyan}{\textbf{-0.23}} & \textcolor{cyan}{\textbf{-0.34}} & \textcolor{cyan}{\textbf{-0.06}} & \textcolor{cyan}{\textbf{-1.29}} & \textcolor{blue}{\textbf{3.54}} & \textcolor{blue}{\textbf{0.64}} \\
\midrule[1pt]
ALBERT\textsubscript{$MS$}           & 31.04    & 36.72  & 35.19  & 34.63  & 33.92  & 59.86  & 36.14 & - & - & - & - & 38.21  \\
ALBERT\textsubscript{$US$}  &  32.77& 38.52& 36.38& 36.00& 37.36& 70.22& 39.53& - & - & - & - & 41.54  \\
$\delta$ & \textcolor{blue}{\textbf{1.73}} & \textcolor{blue}{\textbf{1.8}} & \textcolor{blue}{\textbf{1.19}} & \textcolor{blue}{\textbf{1.37}} & \textcolor{blue}{\textbf{3.44}} & \textcolor{blue}{\textbf{10.36}} & \textcolor{blue}{\textbf{3.39}} & - & - & - & - & \textcolor{blue}{\textbf{3.33}}  \\
\bottomrule[2pt]
\multicolumn{13}{l}{\textcolor{cyan}{cyan} indicates multi-script is better than uni-script models and \textcolor{blue}{blue} indicates vice versa}
\end{tabular}
% }
\end{adjustbox}
\caption{Test accuracy on CSQA}
\label{table-res-csqa}
\end{center}
\end{table*}

\textbf{Article Genre, Sentiment \& Discourse Mode Classification:} We evaluate the models on various other sequence classification datasets that are part of the IndicGLUE benchmark. Here again the uni-script model usually performs better than the multi-script model. However for two tasks in Malayalam (ml) and Tamil (ta) we see better performance for the multi-script model. We already mentioned that there is some tokenization issue for Malayalam which can explain the results for Malayalam. The results for Tamil suggests that it may be a good idea to try both uni-script and multi-script model if they are available to see which performs best on a particular task. However this is the only instance of a task where we see the multi-script model perform better.

\subsection{Zero Shot Capability Testing}
We use the CSQA task to test the zero-shot capability of the models as we can use the models without finetuning.
This task is designed to test whether language models can be used as knowledge bases \citep{petroni-etal-2019-language}. In Table~\ref{table-res-csqa} we report the results. 
We note that the RemBERT models perform much better than the ALBERT models on this task. This is expected as the ALBERT models' memorization capability is hampered by weight sharing.

The ALBERT uni-script model is better on all languages compared to the ALBERT multi-script model. This shows the potential of a uni-script model in a restricted low parameter situation. For the RemBERT models, the results are mixed. However, on average the uni-script model performs better than the  multi-script model. The worst results are for Malayalam (ml) which as we mentiond before has some tokenization issues.
\section{Cross-lingual Representation Similarity}
\label{CLRS}
\begin{figure*}[!htbp]
\begin{center}
\subfloat[ALBERT\textsubscript{$MS$} ]{\includegraphics[width=0.4\textwidth]{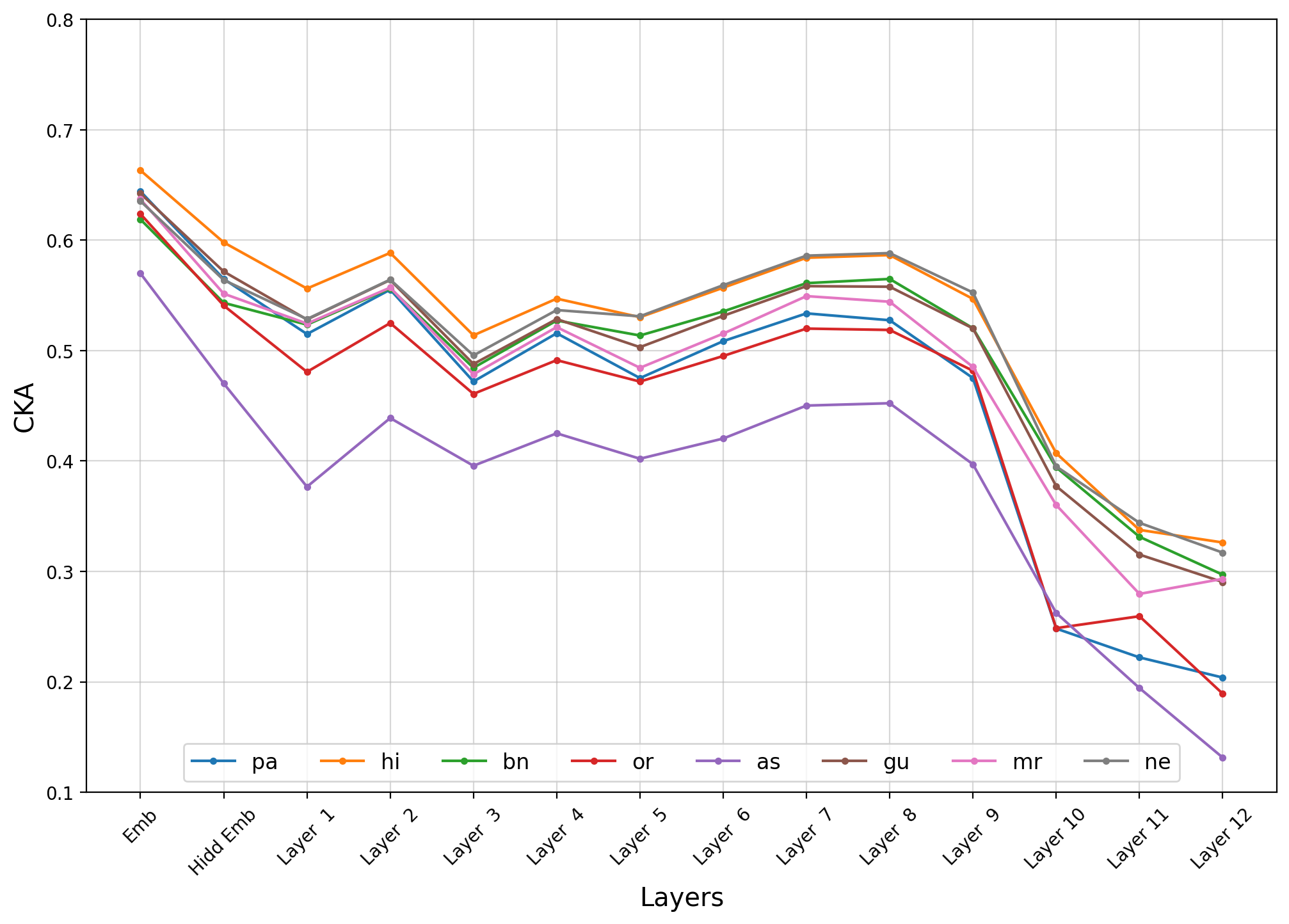}\label{fig:ckaperlangbaselinealbertms}}
\hfill
\subfloat[ALBERT\textsubscript{$US$} ]{\includegraphics[width=0.4\textwidth]{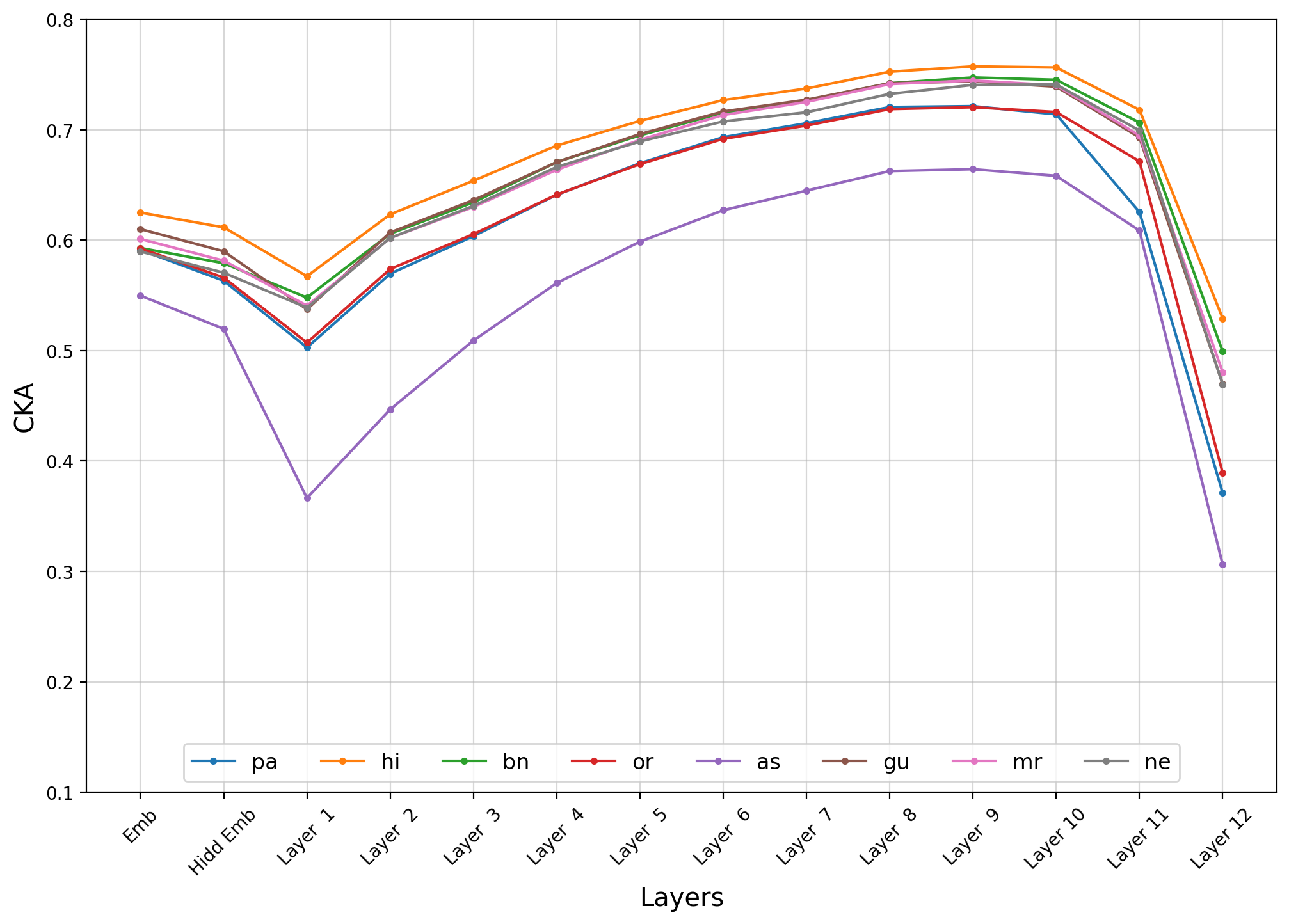}\label{fig:ckaperlangxlmindicalbertus}}
\vfill
\subfloat[RemBERT\textsubscript{$MS$} ]{\includegraphics[width=0.4\textwidth]{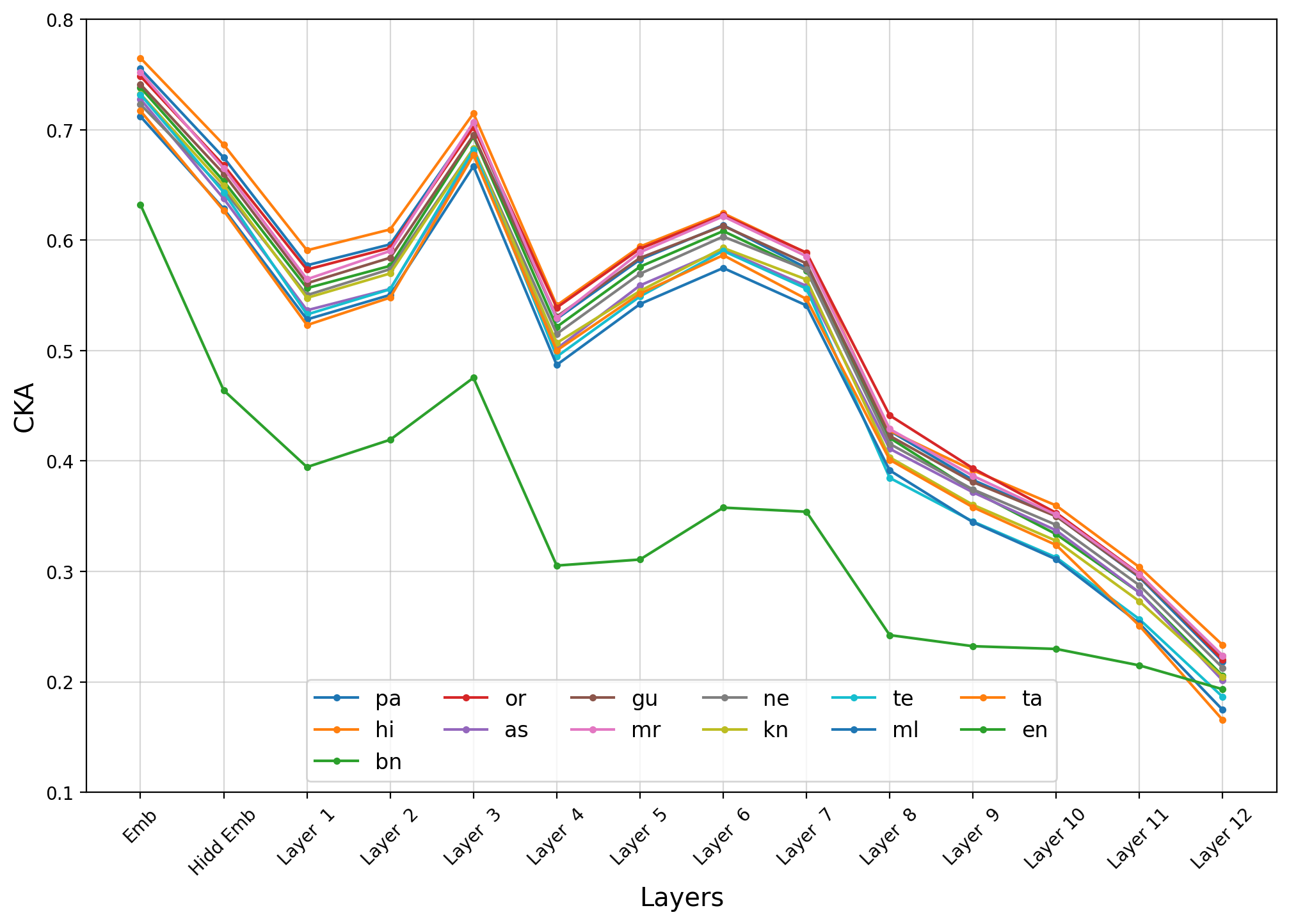}\label{fig:ckaperlangbaselinerembertms}}
\hfill
\subfloat[RemBERT\textsubscript{$US$} ]{\includegraphics[width=0.4\textwidth]{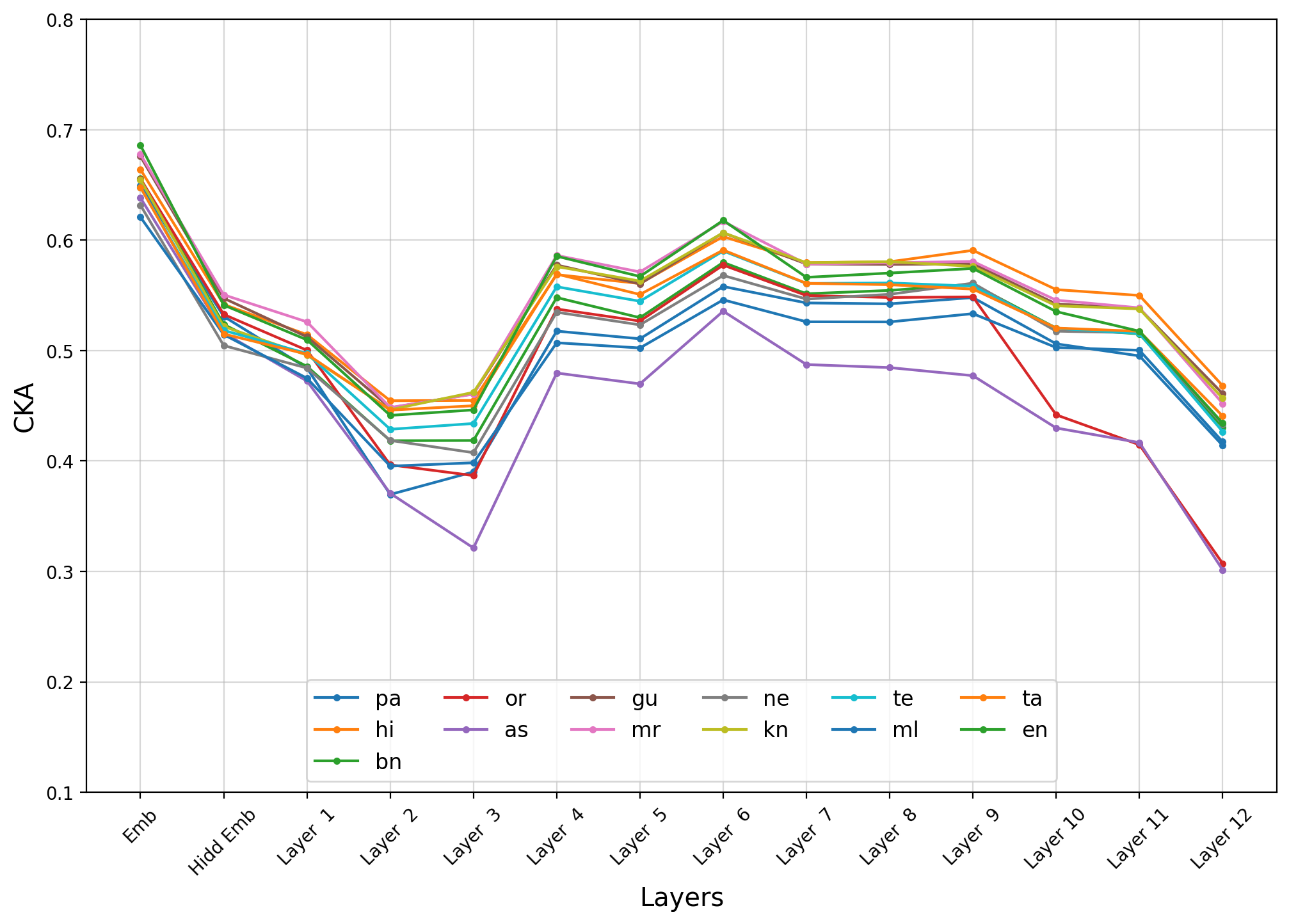}\label{fig:ckaperlangxlmindicrembertus}}
\caption{CKA Similarity Score for the multi-script and uni-script models}
\label{fig:ckaperlang}
\end{center}
\end{figure*}
In this section, we analyze why the uni-script model performs better than the multi-script model from the perspective of Cross-Lingual Representation Similarity.
Following \citep{Muller2021FirstAT}, \citep{Wu2020EmergingCS} and \citep{Del2021EstablishingII} we apply CKA to measure CLRS. We use the CKA implementation from the Ecco library \citep{alammar-2021-ecco}. We use parallel sentences on thirteen languages from the FLORES-101 \citep{Goyal2021TheFE} dataset. For the ALBERT models, which are trained on only the Indo-Aryan languages, we only consider Panjabi, Hindi, Bengali, Oriya, Assamese, Gujarati, Marathi, and Nepali sentences. For the RemBERT models, we additionally consider Kannada, Telugu, Malayalam, Tamil, and English sentences.

First, we calculate the sentence embeddings of these parallel sentences from the models. Sentence embedding is calculated by averaging the hidden state representations of the tokens. Then, we calculate the CKA similarity score between the sentence embeddings for each language pair. For each language, we average its CKA similarity scores. In Figure~\ref{fig:ckaperlang} we plot this average CKA similarity for each layer of the models.

We see that CLRS score drops significantly at the last layer for all models. However, the uni-script models retain high CLRS score until the eleventh layer, whereas the multi-script models have low CLRS score from the ninth layer.
Overall the CLRS score of the uni-script models are more stable. This indicates that the uni-script models have learned better cross-lingual representations.
% We show the individual similarity figures for all language pairs in %appendix~\ref{sec:app_cka}.
\section{Tokenizer Quality Analysis}
 In terms of performance, we expect the transliteration model to exploit better tokenization across the languages. 
 %Whereas the same word written in six different scripts gets mapped to six different tokens in the multi-script model, these get mapped to the same token in the transliteration model. Thus we expect the tokenizer to more efficiently tokenize a given text in the case of the transliteration model..
%Closely related languages often share a lot of vocabulary. Multilingual languages can pick up on these shared words and learn cross-lingual word representations. Pivoting on these shared words, the model can learn the connection between other words. Often a word is rarely used in one language but is frequently used in another language. Multilingual models can learn good representation for the word from the corpus of the later language and improve performance on the former language. Even if the shared words only partially match their spelling, subword tokenizers can often exploit the subword similarity.
\begin{figure*}[hbt!]
\begin{center}
\subfloat[Subword Fertility.]{\includegraphics[width=0.4\textwidth]{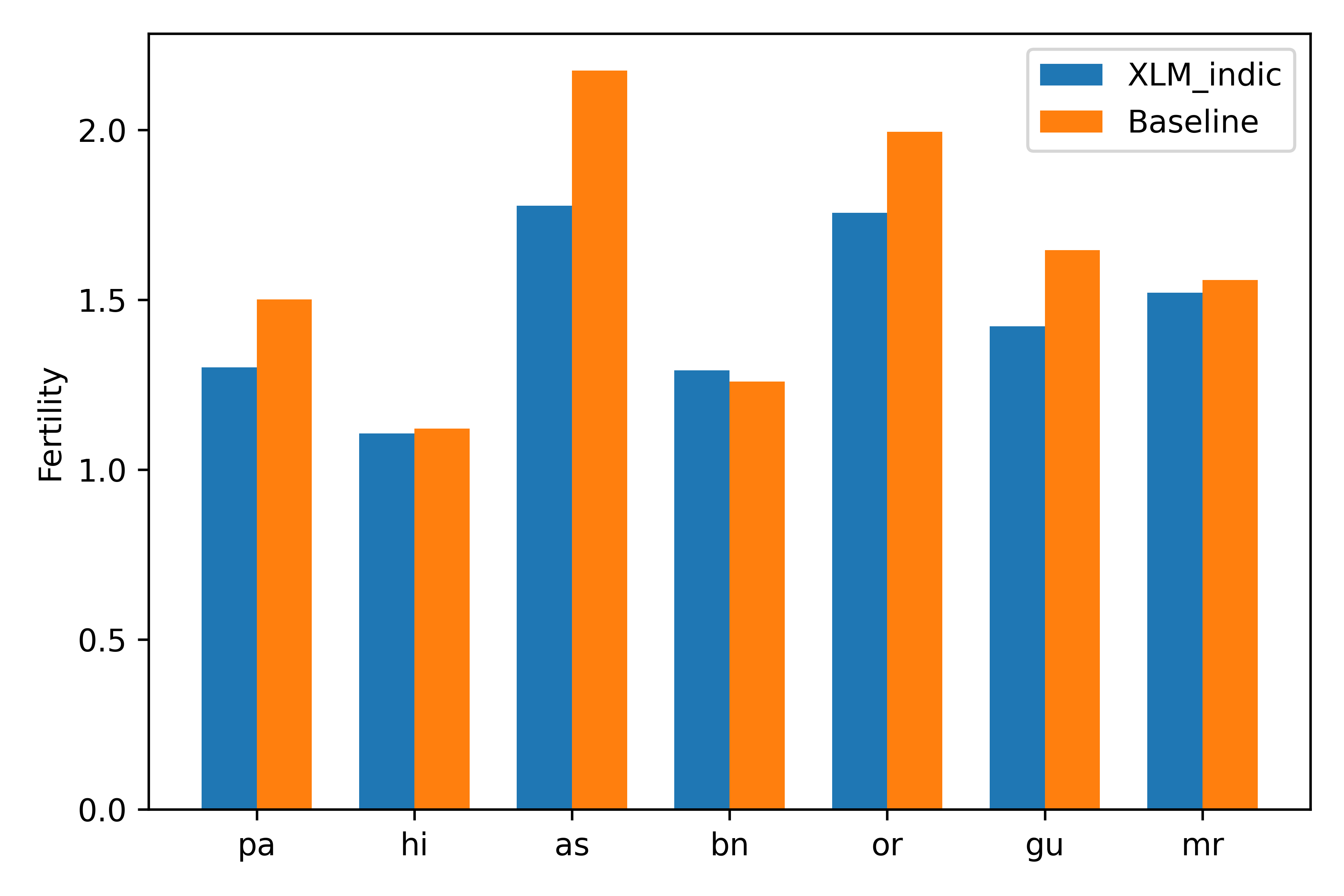}\label{fig:f1}}
\hfill
\subfloat[Unbroken Ratio.]{\includegraphics[width=0.4\textwidth]{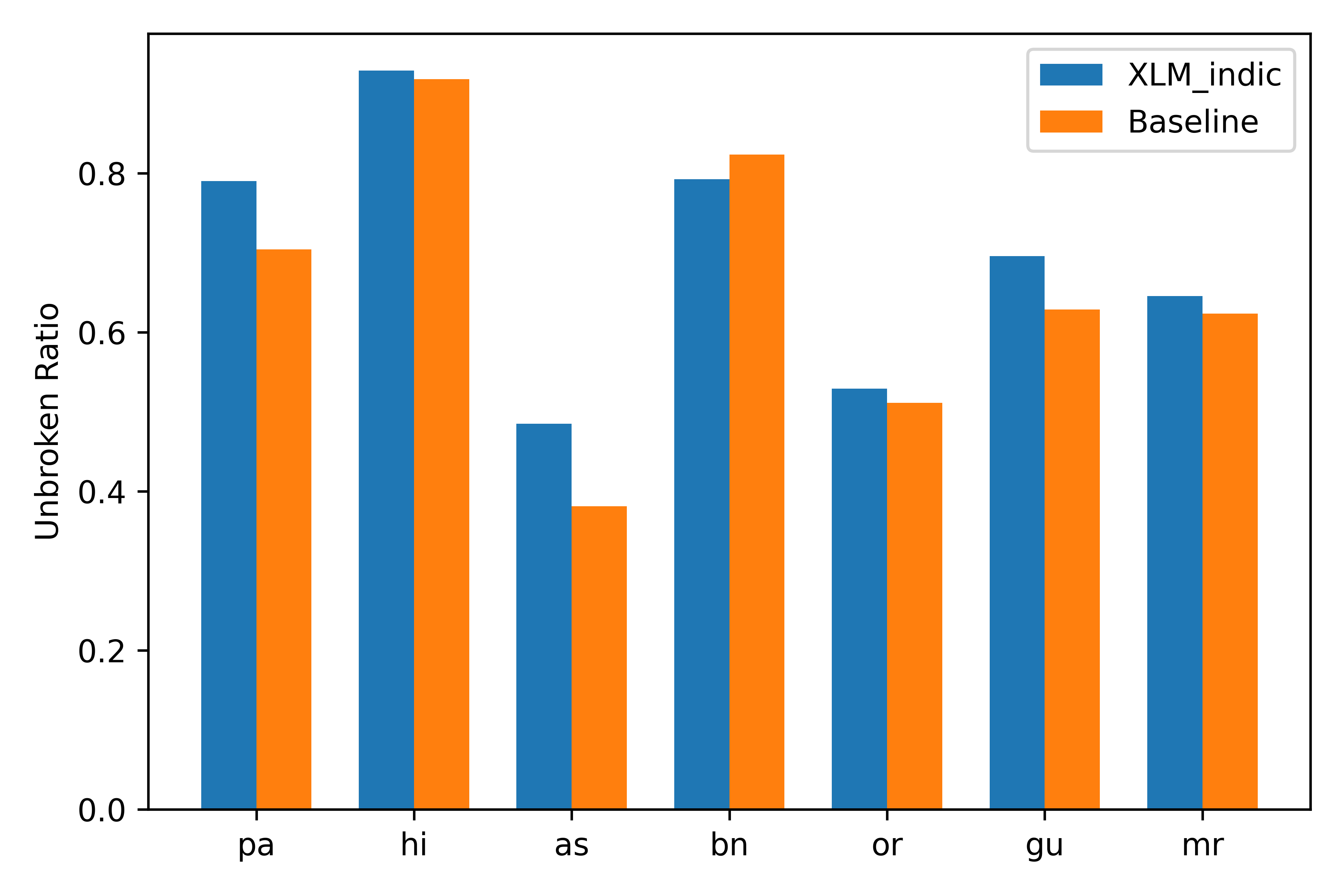}\label{fig:f2}}
\caption{Subword fertility (lower is better) and unbroken ratio (higher is better)\label{tokenizer-stat}  }
\end{center}
\end{figure*}
Following \citep{acs:2019} and \citep{rust-etal-2021-good}, we measure the subword fertility (average number of tokens per word) and the ratio of words unbroken by the tokenizer. From figure~\ref{tokenizer-stat}, we can see that transliteration reduces the splitting of words. This indicates that many words that were represented by different tokens in the multi-script model are represented by a single token in the transliteration model. On average, the ALBERT uni-script tokenizer has a lower subword fertility score of 1.55 compared to the multi-script tokenizer's 1.825. The uni-scirpt tokenizer also has a lower proportion of continued word score of 0.36 while the multi-script tokenizer has a score of 0.45.
\section{Conclusion and Future Work}
In this paper, we show that transliterating closely related languages to a common script improves multilingual language model performance and leads to better cross-lingual representations. We conducted rigorous statistical analysis to quantify the significance and effect size of transliteration on downstream task performance. We found that transliteration especially improves performance on comparatively low-resource languages and did not hurt the performance on high-resource languages. This findings are in agreement with \citep{dhamecha-etal-2021-role,muller-etal-2021-unseen}. 
Our results indicate that in other scenarios where closely related languages use different scripts, transliteration can be used to improve the performance of language models. For example, Slavic and Turkic languages present similar scenarios. 
We would like to extend our study to models at different scales and more languages in the future. Also, another interesting future direction would be to just use the transliteration for pretraining signal but give the model the ability to deal with the original scripts. %Also, character-based models hold more potential to exploit the lexical overlap enabled by transliteration. %Efficient Transformer models that can efficiently handle long sequences can be used to train these character-level models.
\section*{Limitations}
A limitation of our work is that it introduces a transliteration step into the model pipeline. Thus we need a stable implementation of the transliteration scheme. Thus the model can become tied to a specific version of the transliteration library. Also the transliteration scheme is not perfect as we saw for Malayalam, it introduced some artifacts. Finally given our limited computational budget, we could not run experiments with a lot of models at different scales. Thus the impact of transliteration over different model scales has not been explored. Even though our work has these limitations, it clearly shows transliteration as an important tool for training better multilingual models.
\section*{Ethics Statement}
%In this paper, we empirically show the effect of transliteration on the performance of Multilingual Language Models. 
In their study, \citet{joshi-etal-2020-state} showed the resource disparity between low-resource and high-resource languages, and \citep{ruder2020beyondenglish} also highlighted the necessity of working with low-resource languages. However, creating representative and inclusive corpora is a difficult task and an ongoing process and is not always possible for many low-resource languages. Thus to inclusively advance the state of NLP across languages, it is crucial to develop techniques for training MLLMs that can extract the most out of existing multilingual corpora. Hence, we believe our analysis might help MLLMS with low-resource languages in real-world applications. However, there is one ethical issue that we want to state explicitly. Even though we pretrain on a comparatively large multilingual corpus, the model may exhibit harmful gender, ethnic and political bias. If the model is fine-tuned on a task where these issues are important, it is necessary to take special consideration when relying on the model's decisions.
\bibliographystyle{acl_natbib}
\bibliography{custom}

\appendix
\section{Cloze Style QA Evaluation Method}
\label{sec:csqa-eval-isse}
Since a word can be tokenized to multiple tokens by the subword tokenizer, correctly evaluating the model on this task requires special care. Specifically, we have to use the same number of mask tokens as the number of subword tokens that a word gets split into. Then we calculate the probability for the word by multiplying the probability of the subword tokens  predicted by the masked language model. %We found that on the IndicBERT evaluation code, only a single mask token was used irrespective of the number of subword tokens that a word gets split into. We do not think this is the correct way to evaluate a masked language model on this task.

\begin{table*}[hbt!]
\begin{center}
\begin{adjustbox}{max width=\textwidth}
\small
\begin{tabular}{l c c c c c c c}
\\ \hline \\
\textbf{Task} & \textbf{TPU} & \textbf{Batch Size} & \textbf{Learning Rate} & \textbf{Weight Decay} & \textbf{Dropout} & \textbf{Epochs} & \textbf{Warmup Ratio} \\
\\ \hline \\
News Category Classification  & False & 16 & 2e-5 & 0.01 & 0.1 & 20 & 0.10 \\
Wikipedia Section-Title Prediction & True & 256 & 2e-5 & 0.01 & 0.1 & 3 & 0.10 \\
Named Entity Recognition & True & 512 & 2e-5 & 0.01 & 0.1 & 20 & 0.10 \\
BBC Hindi News Classification & False & 16 & 2e-5 & 0.01 & 0.1 & 20 & 0.10 \\
Soham Bengali News Classification & False & 16 & 2e-5 & 0.01 & 0.1 & 8 & 0.10 \\
INLTK Headlines Classification & False & 256 & 2e-5 & 0.01 & 0.1 & 20 & 0.10 \\
IITP Movie Review & False & 64 & 5e-5 & 0.01 & 0.25 & 20 & 0.10 \\
IITP Product Review & False & 16 & 5e-5 & 0.01 & 0.5 & 20 & 0.10 \\
MIDAS Discourse Mode & False & 32 & 2e-5 & 0.01 & 0.5 & 20 & 0.10 \\
\\ \hline \\
\end{tabular}
\end{adjustbox}
\end{center}
\caption{Hyperparameters for ALBERT models}
\label{downstream-hyperparameters-tb-albert}
\end{table*}

\begin{table*}[hbt!]
\begin{center}
\begin{adjustbox}{max width=\textwidth}
\small
\begin{tabular}{l c c c c c c c}
\\ \hline \\
\textbf{Task} & \textbf{TPU} & \textbf{Batch Size} & \textbf{Learning Rate} & \textbf{Weight Decay} & \textbf{Dropout} & \textbf{Steps} & \textbf{Label Smoothing} \\
\\ \hline \\
News Category Classification  & False & 16 & 1e-5 & 0.1 & 0.1 & 2500 & 0.0 \\
Wikipedia Section-Title Prediction & True & 256 & 8e-6 & 0.1 & 0.1 & 12500 & 0.0 \\
Named Entity Recognition & False & 16 & 5e-5 & 0.1 & 0.1 & 10000 & 0.0 \\
BBC Hindi News Classification & False & 16 & 1e-5 & 0.01 & 0.1 & 2500 & 0.0 \\
Soham Bengali News Classification & False & 16 & 1e-5 & 0.1 & 0.1 & 2500 & 0.1 \\
INLTK Headlines Classification & False & 16 & 1e-5 & 0.1 & 0.1 & 5000 & 0.0 \\
IITP Movie Review & False & 16 & 1e-5 & 0.1 & 0.1 & 5000 & 0.0 \\
IITP Product Review & False & 16 & 1e-5 & 0.1 & 0.1 & 5000 & 0.0 \\
ACTSA Sentiment Classification & False & 16 & 1e-5 & 0.1 & 0.1 & 5000 & 0.0 \\
MIDAS Discourse Mode & False & 16 & 8e-6 & 0.1 & 0.1 & 2500 & 0.1 \\
\\ \hline \\
\end{tabular}
\end{adjustbox}
\end{center}
\caption{Hyperparameters for RemBERT models}
\label{downstream-hyperparameters-tb-rembert}
\end{table*}

\section{Pretraining Details}
\label{sec:pretraining}
\textbf{Corpus Preparation:}
Since the OSCAR corpus contains raw text from the Web, we apply a few filtering and normalization. First, we discard entries where the dominant script does not match the language tag provided by the OSCAR corpus. Then we use the IndicNLP normalizer \citep{kunchukuttan2020indicnlp} to normalize the raw text. For the uni-script model, we then transliterate all the text to ISO-15919 format using the Aksharamukha \citep{Aksharamukha} library.

For the RemBERT models we do not perform any of the filtering mentioned above since our pretraining corpus is comparatively very large. In this case, we use the PyICU library \citep{pyicu} for transliterating to ISO-15919 format.

\textbf{Tokenizer Training:} For the ALBERT models, we train two SentencePiece tokenizers \citep{Kudo2018SentencePieceAS} on the transliterated and the non-transliterated corpus with a vocabulary size of 50,000. For the RemBERT models we train Unigram tokenizers from the Tokenizers library \citep{wolf-etal-2020-transformers} with a vocabulary size of 65,536.
%Then we use the trained tokenizer and the sentence-splitter from the IndicNLP library to split long entries from the corpus at sentence boundaries so that no entry may have more than 512 tokens. Finally, we discard short entries (<512 characters) to improve the training efficiency. 

\textbf{ALBERT Model Training:} We first pretrained an ALBERT base model from scratch on the non-transliterated corpus as our baseline. Afterward, we pretrained another ALBERT base from scratch on the transliterated corpus. We chose the base model due to computing constraints. We trained the models on a single TPUv3 VM. Both models were trained using the same hyperparameters. We followed the hyperparameters used in \citep{Lan2020ALBERTAL} except for batch size and learning rate. The pretraining objective is also the same as \citep{Lan2020ALBERTAL}.We used a batch size of 256, which is the highest that fits into TPU memory, whereas the ALBERT paper used a batch size of 4096. As our batch size is 1/16\textsuperscript{th} of the ALBERT paper, we use a learning rate of 1e-3/8, which is approximately 1/16\textsuperscript{th} of the learning rate used in the ALBERT paper (1.76e-2). Additionally, we use the Adam optimizer \citep{Kingma2015AdamAM} instead of the LAMB optimizer. The rest of the hyperparameters were the same as the ALBERT paper. Specifically, we use a sequence length of 512 with absolute positional encoding, weight decay of 1e-2, warmup steps of 5000, max gradient norm of 1.0, and Adam epsilon of 1e-6. The models were trained for 1M steps. Each model took about 7.5 days to train. We use the ALBERT implementation from the Huggingface Transformers Library \citep{wolf-etal-2020-transformers}. 

\textbf{RemBERT Model Training:} We pretrained an RemBERT base models similar to the ALBERT models. We trained the models on a single TPUv3 VM. Both models were trained using the same hyperparameters. We followed the hyperparameters used in \citep{Chung2021RethinkingEC} except for batch size and learning rate. The pretraining objective is also the same as \citep{Chung2021RethinkingEC}. We used a batch size of 256, which is the highest that fits into TPU memory, whereas the RemBERT paper used a batch size of 2048. As our batch size is 1/8\textsuperscript{th} of the RemBERT paper, we use a learning rate of 2e-4/8, which is  1/8\textsuperscript{th} of the learning rate used in the RemBERT paper. Similar to the ALBERT model, we use the Adam optimizer \citep{Kingma2015AdamAM}. The rest of the hyperparameters were the same as the RemBERT paper. Specifically, we use a sequence length of 512 with absolute positional encoding, weight decay of 1e-2, warmup steps of 15000, max gradient norm of 1.0, and Adam epsilon of 1e-6. The models were trained for 1M steps. Each model took about 7.5 days to train. We use the RemBERT implementation from the Huggingface Transformers Library \citep{wolf-etal-2020-transformers}.
\section{Downstream Hyperparameters}
\label{sec:hype}
Hyperparameters for downstream tasks are presented in Table~\ref{downstream-hyperparameters-tb-albert} and Table~\ref{downstream-hyperparameters-tb-rembert}.

For the ALBERT models batch size was chosen to be the maximum that fits in memory. This was done so that each batch contains approximately the same number of tokens. Otherwise the hyperparameters were chosen following the recommendations of \citep{DBLP:conf/iclr/MosbachAK21}. On the highly skewed IITP Movie Review, IITP Product Review and MIDAS Discourse we found that this default setting resulted in worse performance compared to the independent baselines. So we finetuned the learning rate and classifier dropout on the validation set of these tasks.

For the RemBERT models learning rate, weight decay, dropout, steps and label smoothing were chosen based on grid search with a few values.

\begin{table*}[hbt!]
\begin{center}
\begin{adjustbox}{max width=\textwidth}
\resizebox{0.7\textwidth}{!}{
\small
\begin{tabular}{l c c c c c c c c c c c}
\toprule[2pt]
\textbf{Model} & \textbf{pa} & \textbf{hi} & \textbf{bn} & \textbf{or} & \textbf{as} & \textbf{gu} & \textbf{mr} & \textbf{kn} & \textbf{te} & \textbf{ml} & \textbf{ta}\\
\midrule[2pt]
\multicolumn{12}{l}{\textbf{Wikipedia Section Title Prediction}}\\
RemBERT\textsubscript{$\rho$}   & 1 & 1 & 1 & 1 & 1 & 1 & 0.91 & 1 & 1 & 0.67 & 0.99\\
RemBERT\textsubscript{$r$}   & 0.83 & 0.83 & 0.83 & 0.84 & 0.83 & 0.84 & 0.69 & 0.83 & 0.83 & 0.27 & 0.81 \\
\midrule[1pt]
ALBERT\textsubscript{$\rho$}           &1 &1 &1	&1	&1	&1	&1 &- & - &- & -\\
ALBERT\textsubscript{$r$}   & 0.83 & 0.83 & 0.83 & 0.83 & 0.83 & 0.83 & 0.83 & - &- & - & -\\
\midrule[2pt]
\multicolumn{12}{l}{\textbf{News Category Classification}}\\
RemBERT\textsubscript{$\rho$}   & 1 & - & 0.27 & 1 & - & 0.87 & 0.46 & 0.27 & 0.45 & 0.94 & 0.75 \\
RemBERT\textsubscript{$r$}   & 0.85 & - & 0.39 & 0.84 & - & 0.68 & 0.07 & 0.39 & 0.07 & 0.75 & 0.41 \\
\midrule[1pt]
ALBERT\textsubscript{$\rho$}    &1 &-	&0.31	&1	&-	&0.80	&0.31 & - &- & - & -  \\
ALBERT\textsubscript{$r$}   & 0.86 & - & 0.32 & 0.83 & - & 0.51 & 0.32 & - &- & - & - \\
\midrule[2pt]
\multicolumn{12}{l}{\textbf{\textbf{Named Entity Recognition}}}\\
RemBERT\textsubscript{$\rho$} & 1.00 & 0.95 & 0.99 & 0.70 & 0.91 & 1.00 & 0.57 & 1.00 & 1.00 & 0.56 & 0.65\\
RemBERT\textsubscript{$r$} & 0.83 & 0.75 & 0.81 & 0.33 & 0.69 & 0.83 & 0.10 & 0.83 & 0.83 & 0.08 & 0.25 \\
\midrule[1pt]
ALBERT\textsubscript{$\rho$}   &1 &1	&1	&1	&1	&1	&1 & - &- & - & -\\
ALBERT\textsubscript{$r$}   & 0.83 & 0.83 & 0.83 & 0.83 & 0.83 & 0.83 & 0.83 & - &- & - & -\\
\bottomrule[2pt] 
\end{tabular}
}
\end{adjustbox}
\end{center}
\caption{Test Statistics on Classification Tasks from IndicGLUE Benchmark}
\label{table-indglue-rho&stdeffect}
\end{table*}
% The MWU p-values and test statistics for public datasets accuracy is given in Table~\ref{table-mw-pb}. We can see that for BBC News (p-value 0.0088) and Soham Articles Classification (p-value 0.0090) the uni-script model is better than the multi-script model with a $\delta$ of 1.86 and 0.67, respectively. Both tasks have large $r$  (0.62 for both). However, as per $\rho$, the uni-script model outperforms the multi-script model 87\% of the time.  Whereas, the $\rho$ for Soham News Article Classification is 0.57. As for INLTK Headlines, the uni-script model and the multi-script model perform equivalently. On INLTK Headlines, the p-value for Gujarati ($\delta$ of 0.32) is 0.6249 and Marathi ($\delta$ of -0.17) is 0.3503. On IITP Product Reviews, the uni-script model outperforms the multi-script model with a $\delta$ of 0.85, p-value of 0.4099 and $\rho$ of 0.79.However, the $r$  is medium (0.48) for the task. In contrast, on IITP Movie Reviews, both models are equivalent performance wise with a $\delta$ of 0.15, p-value of 0.8941, $\rho$ of 0.52 and a small (0.031) standardized effect. Finally, we can see that both models performing equally on Discourse Mode Classification. The $\delta$ is 0.15 with a p-value of 0.7561 and a small (0.073) $r$ . However, as per the $\rho$, the uni-script model outperforms the multi-script model 45\% of the time. 
\begin{table*}[hbt!]
\centering
%\begin{adjustbox}{0.8\textwidth}
\resizebox{0.8\textwidth}{!}{
\small
\begin{tabular}{l c| c c| c c}
\toprule[2pt]
%\textbf{Language} & \textbf{Dataset} &\multicolumn{2}{c}{\textbf{multi-script}} &\multicolumn{2}{c}{\textbf{uni-script}} &\multicolumn{2}{c}{\textbf{Test Statistics}} \\
\textbf{Language} & \textbf{Dataset}  &RemBERT\textsubscript{$\rho$} & RemBERT\textsubscript{$r$} &ALBERT\textsubscript{$\rho$}           &ALBERT\textsubscript{$r$} \\
\midrule[2pt]
\multicolumn{6}{l}{\textbf{Article Genre Classification}}\\
hi          & BBC News  & 0.78 & 0.47 & 0.87 & 0.62\\
bn          & Soham News Article Classification & 1 &0.84 & 0.87 & 0.62\\
gu          & INLTK Headlines  & 1 &0.84  & 0.57 & 0.12\\
mr          & INLTK Headlines  & 0.98 &0.79  & 0.36 & 0.22\\
ml          & INLTK Headlines  & 0.08 &0.70 &- &- \\
ta          & INLTK Headlines  & 0.15 &0.59 &- &- \\
\midrule[2pt] 
\multicolumn{6}{l}{\textbf{Sentiment Analysis}}\\
hi          & IITP Product Reviews  & 0.51 &0.01     & 0.79 & 0.48 \\
hi          & IITP Movie Reviews & 0.93 &0.72 & 0.52 & 0.03  \\
te          & ACTSA & 0.31 &0.30 &- &- \\
\midrule[2pt]
\multicolumn{6}{l}{\textbf{Discourse Mode Classification}}\\
hi          & MIDAS Discourse & 0.79 &0.48 & 0.45 & 0.07  \\
\bottomrule[2pt]
\end{tabular}
}
\caption{Test Statistics on Public Datasets}
\label{table-public-rho&stdeffect}
%\end{adjustbox}
\end{table*}

\section{Test Statistics Results}
\label{sec:rho&stdeffect}

$\rho$ gives us the probability of one group being better than the other group. That is the probability that a random performance sample of the the uni-script model is greater than a random performance sample of the multi-script  model. The last test statistic is $r$ which indicates the magnitude of difference between the performance values of the uni-script model (group 1) and the multi-script model (group 2). $r$ shows us how realistically significant the performance differences are between models even if the performance difference is statistically significant. It gives us a value between 0 to 1 and its ranges are: \textbf{small effect}  ( 0 $\leq$ $r$ $\leq$ 0.3) , \textbf{medium effect} ( 0.3 $<$ $r$ $\leq$ 0.5) and \textbf{large effect} (0.5 $<$ $r$). We performed MWU on all downstream tasks except CSQA. On CSQA, we only report the $\delta$. The MWU is performed using the SciPy library \citep{2020SciPy-NMeth}, and the results are further validated using R \citep{sjstats}. These statistic are reported in Table~\ref{table-indglue-rho&stdeffect} for the IndicGLUE classification tasks and in Table~\ref{table-public-rho&stdeffect} for the public dataset classification tasks.

\section{Cross-lingual Similarity of ALBERT Models on All Language Pairs}
\label{sec:app_cka}
\begin{figure*}[hbt!]
\begin{center}
\centering
\subfloat[multi-script PA-X]{\includegraphics[width=0.40\textwidth]{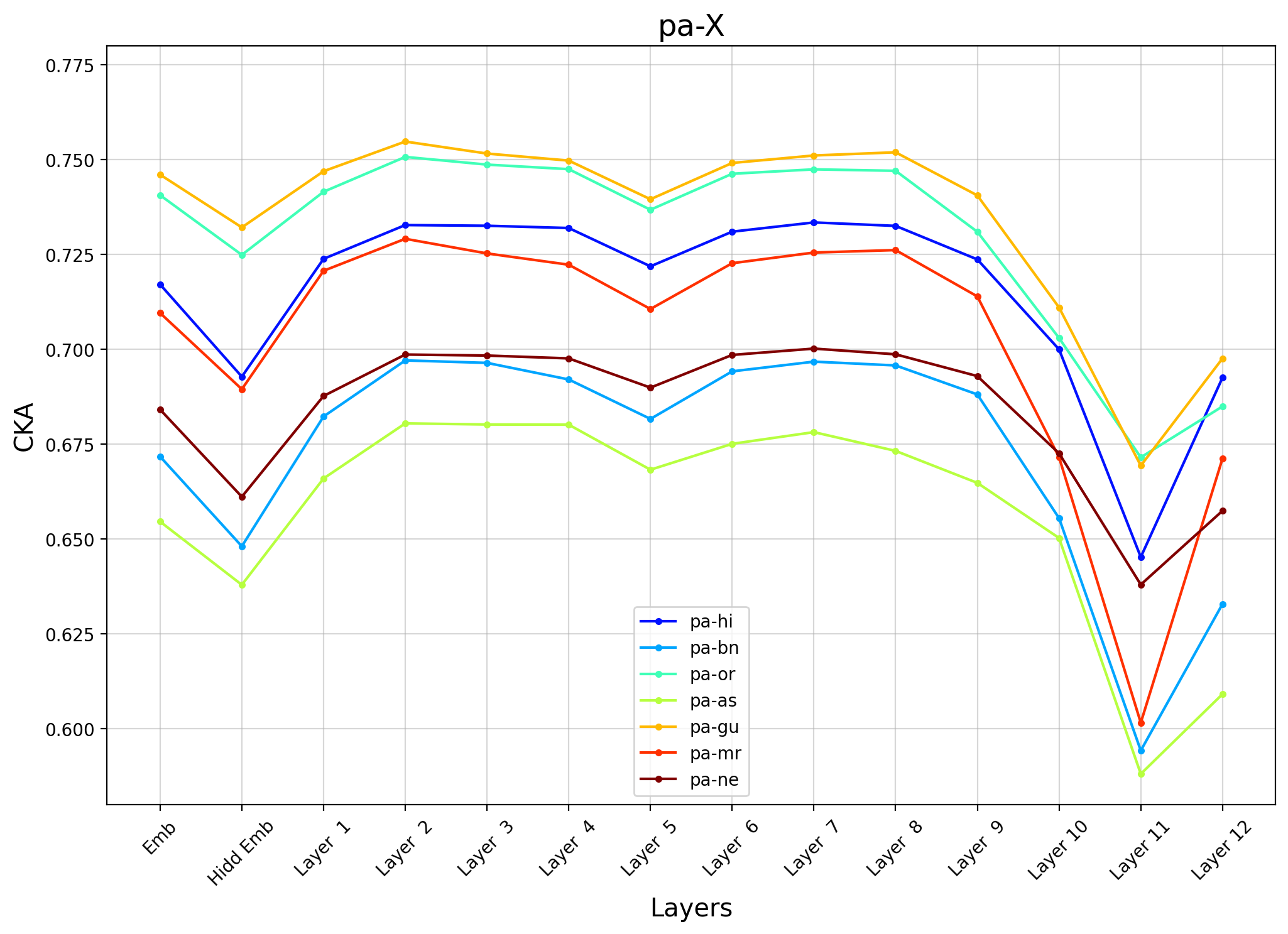}\label{fig:ckaperlangbaseline_pa}}
\hfill
\subfloat[uni-script PA-X]{\includegraphics[width=0.40\textwidth]{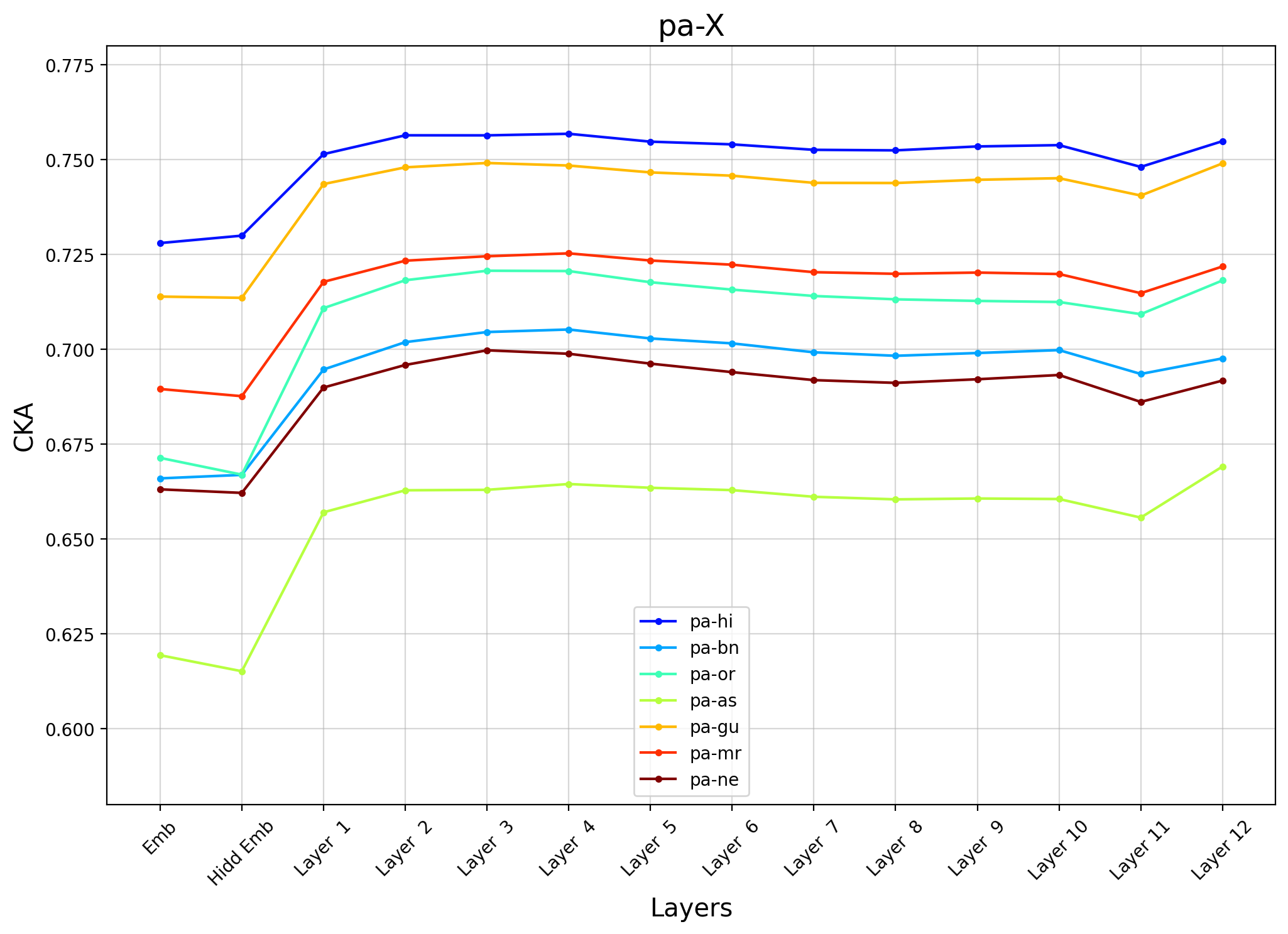}\label{fig:ckaperlangxlmindic_pa}}
\hfill
\subfloat[multi-script HI-X]{\includegraphics[width=0.40\textwidth]{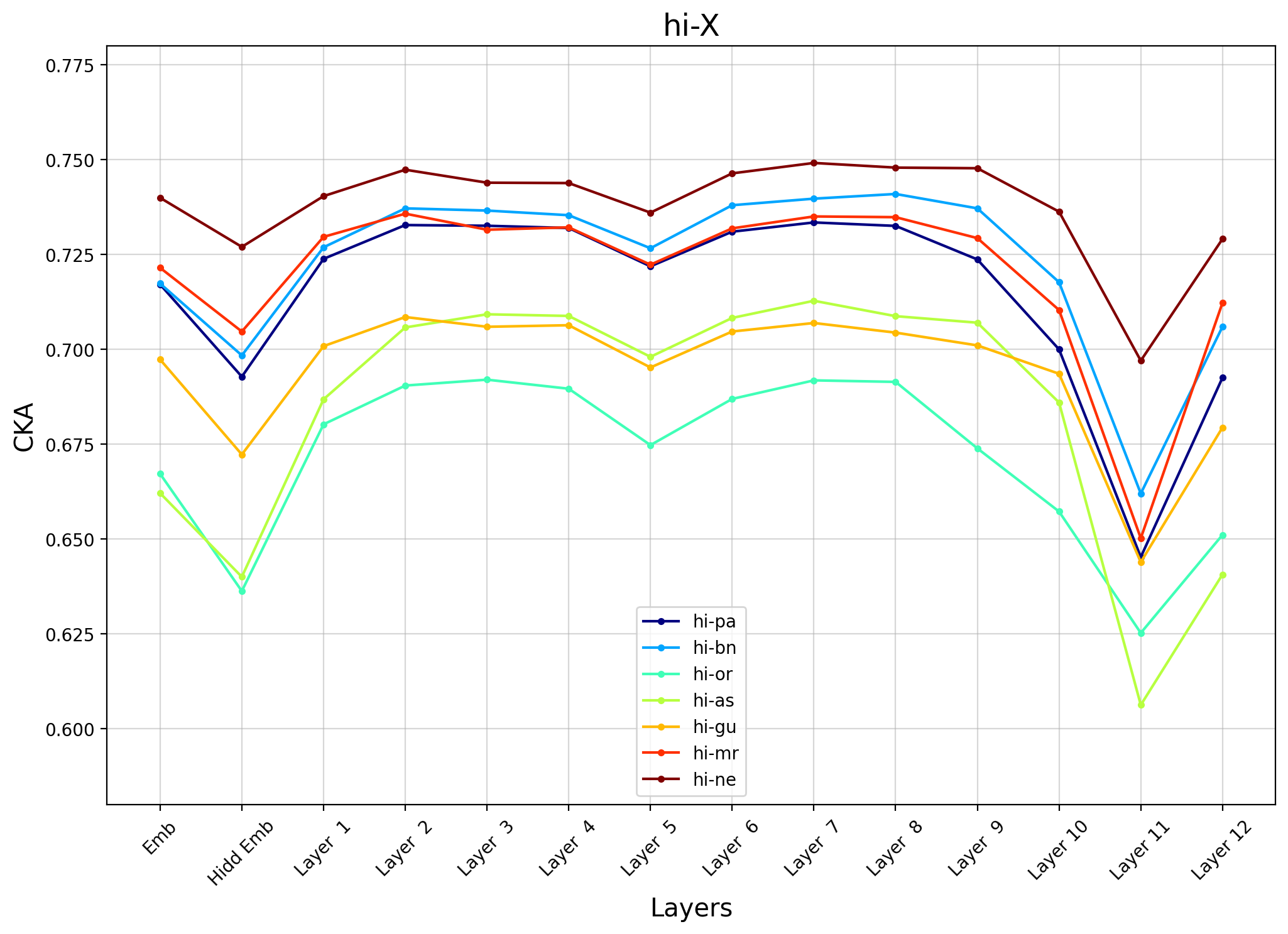}\label{fig:ckaperlangbaseline_hi}}
\hfill
\subfloat[uni-script HI-X]{\includegraphics[width=0.40\textwidth]{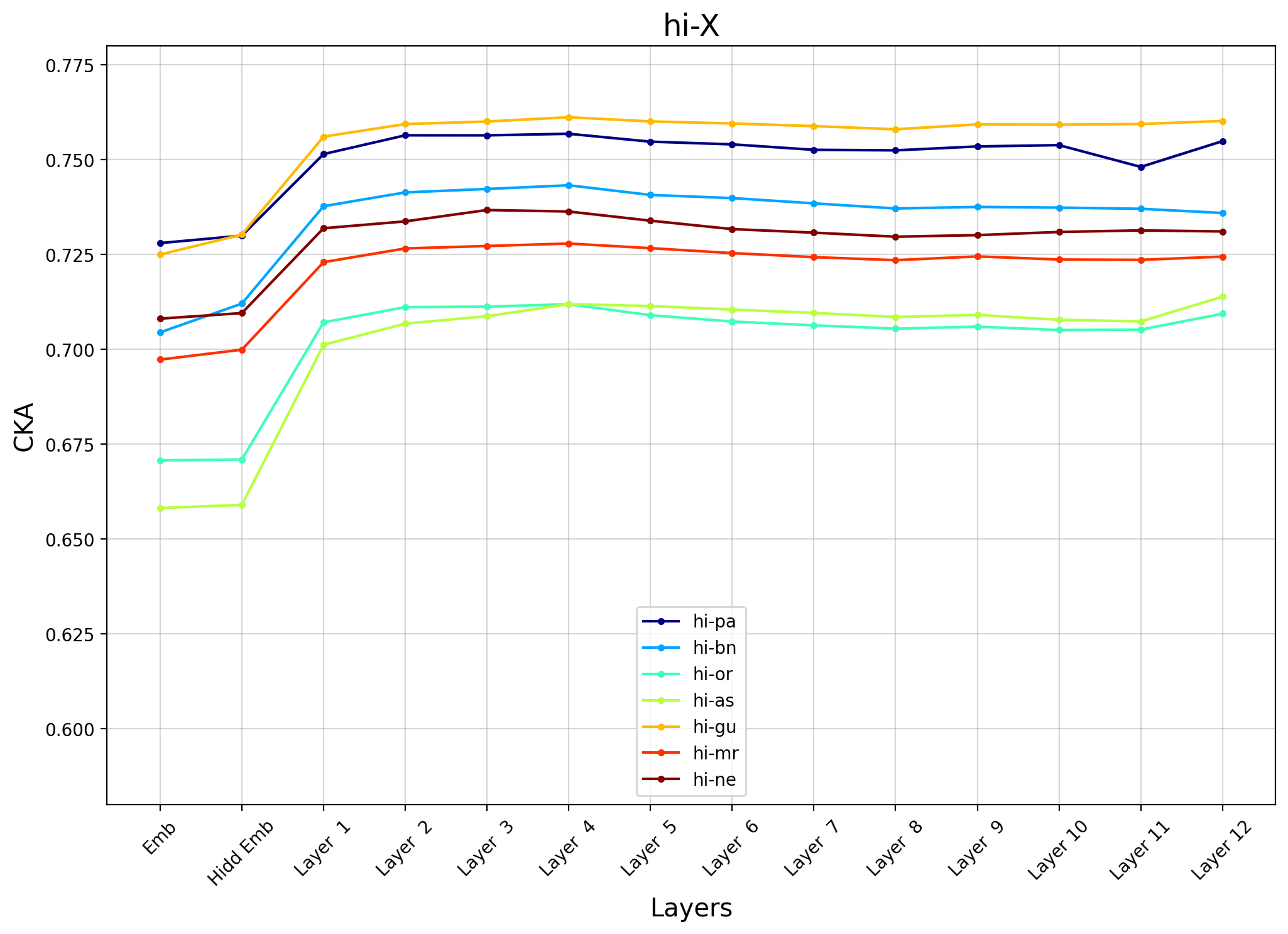}\label{fig:ckaperlangxlmindic_hi}}
\hfill
\subfloat[multi-script BN-X]{\includegraphics[width=0.40\textwidth]{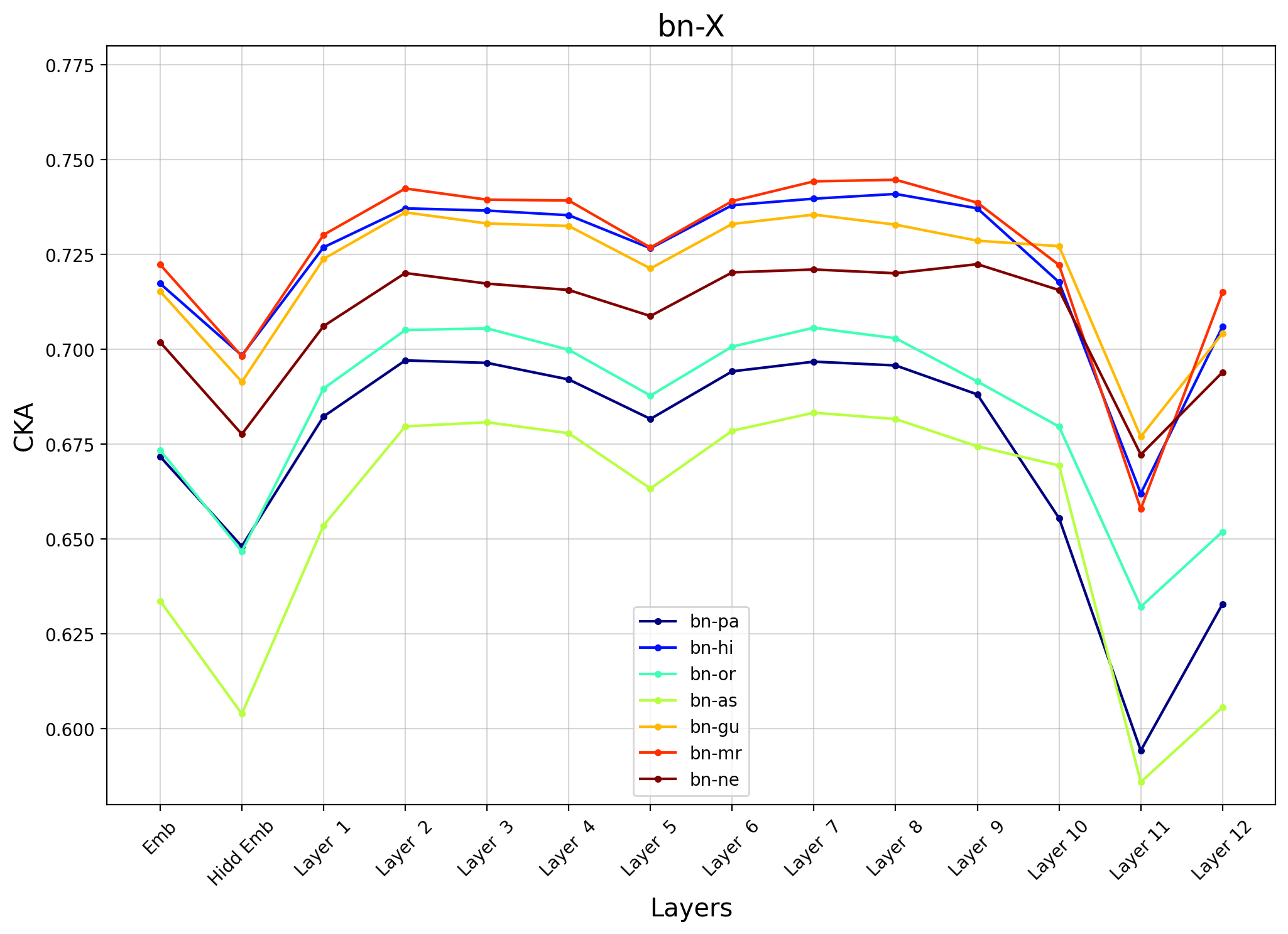}\label{fig:ckaperlangbaseline_bn}}
\hfill
\subfloat[uni-script BN-X]{\includegraphics[width=0.40\textwidth]{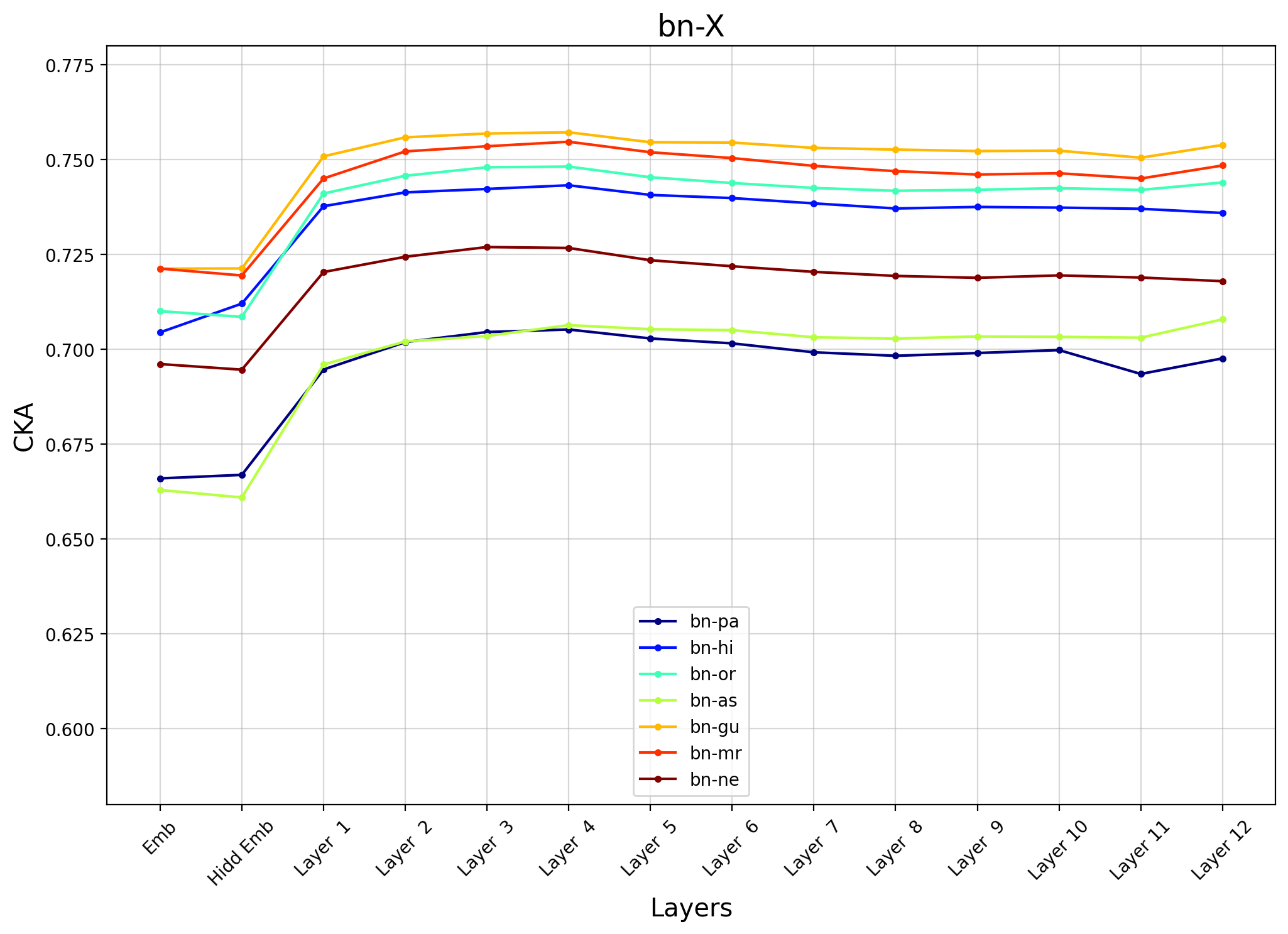}\label{fig:ckaperlangxlmindic_bn}}
\hfill
\subfloat[multi-script OR-X]{\includegraphics[width=0.40\textwidth]{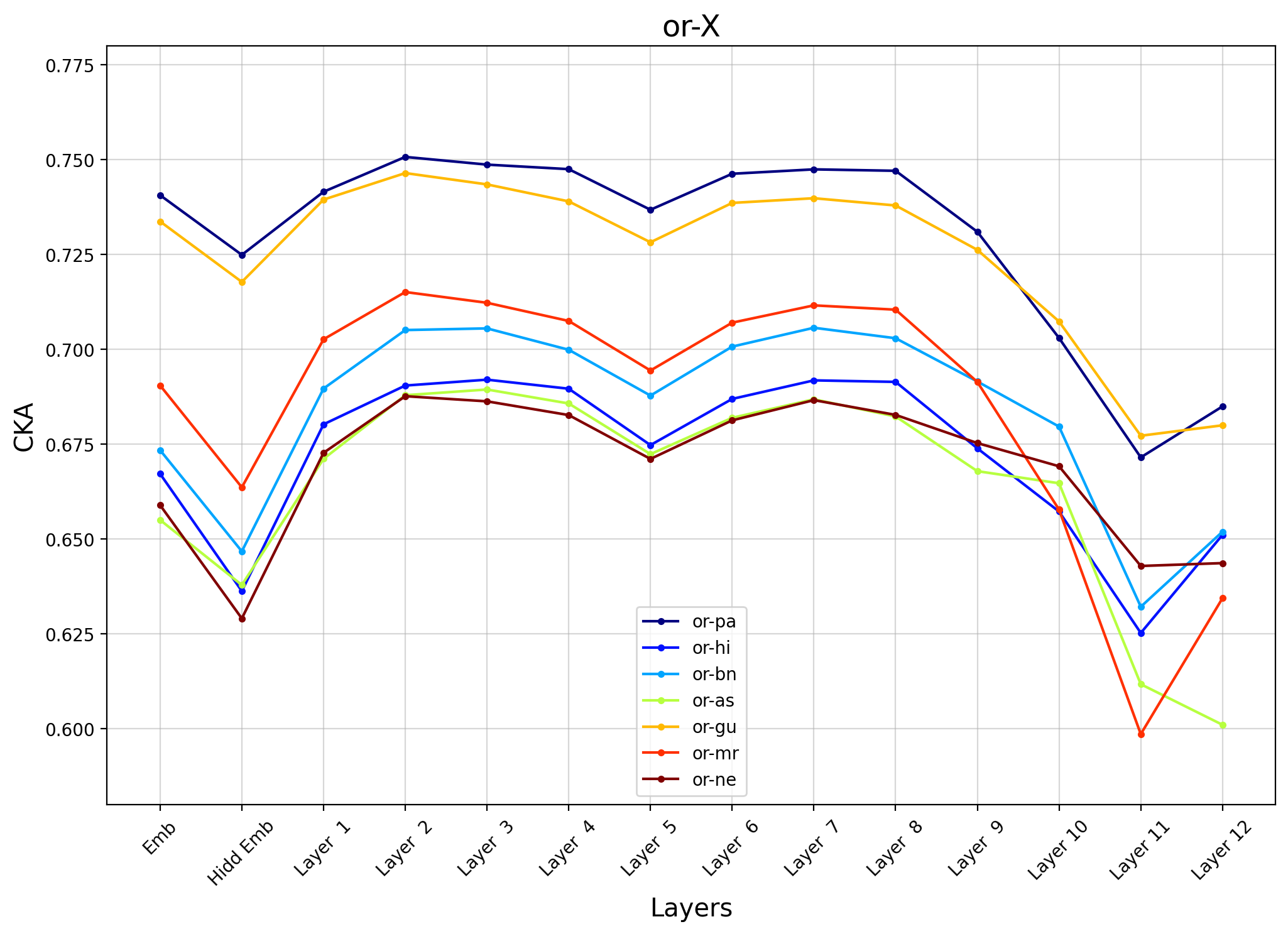}\label{fig:ckaperlangbaseline_or}}
\hfill
\subfloat[uni-script OR-X]{\includegraphics[width=0.40\textwidth]{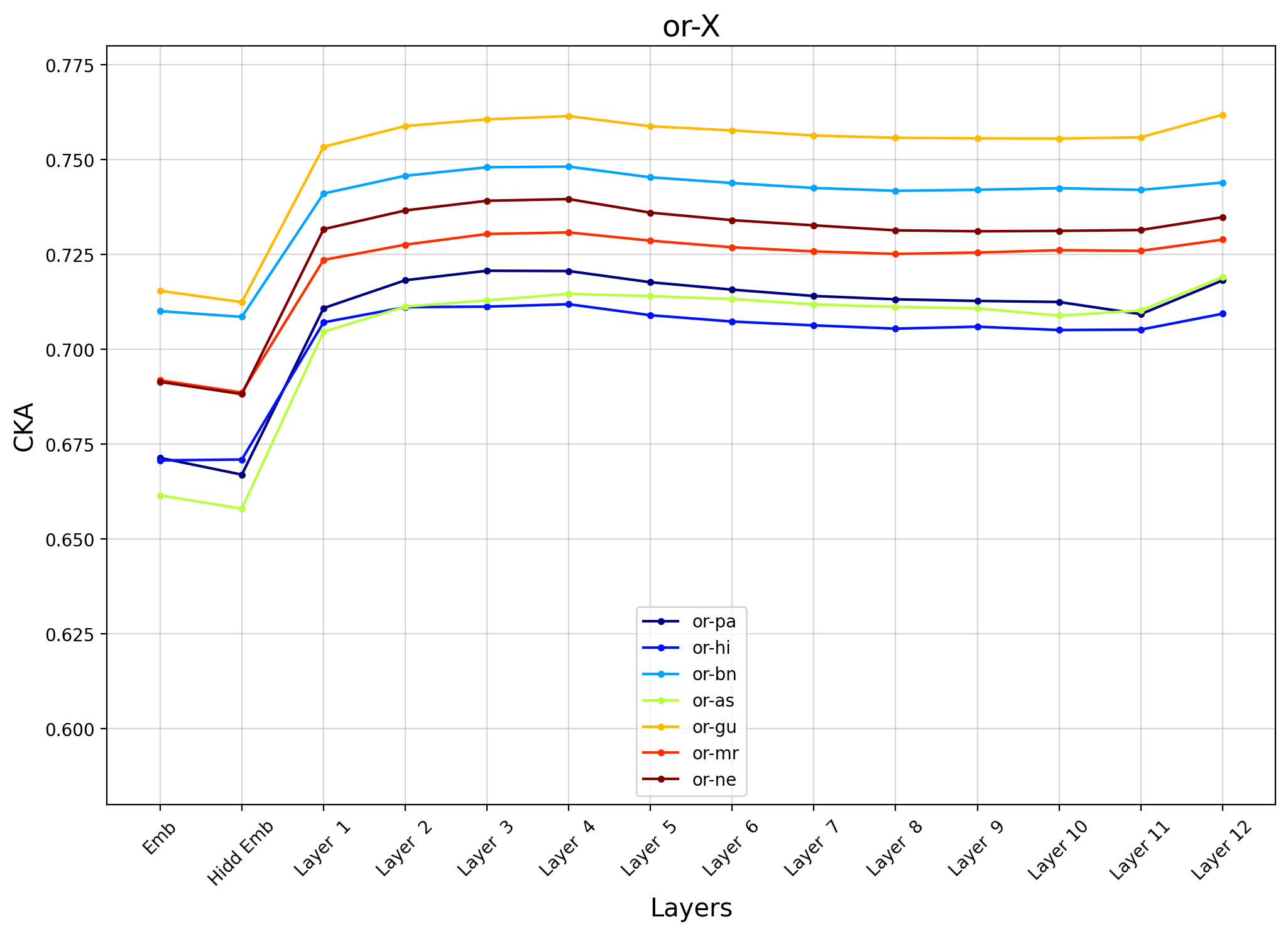}\label{fig:ckaperlangxlmindic_or}}
\hfill

\caption{CKA of multi-script and uni-script on all language pairs for pa, hi,bn and or}
\label{cka_all_language_pairs_pa_to_or}
\end{center}
\end{figure*}

\begin{figure*}[hbt!]
\begin{center}
\subfloat[multi-script AS-X]{\includegraphics[width=0.40\textwidth]{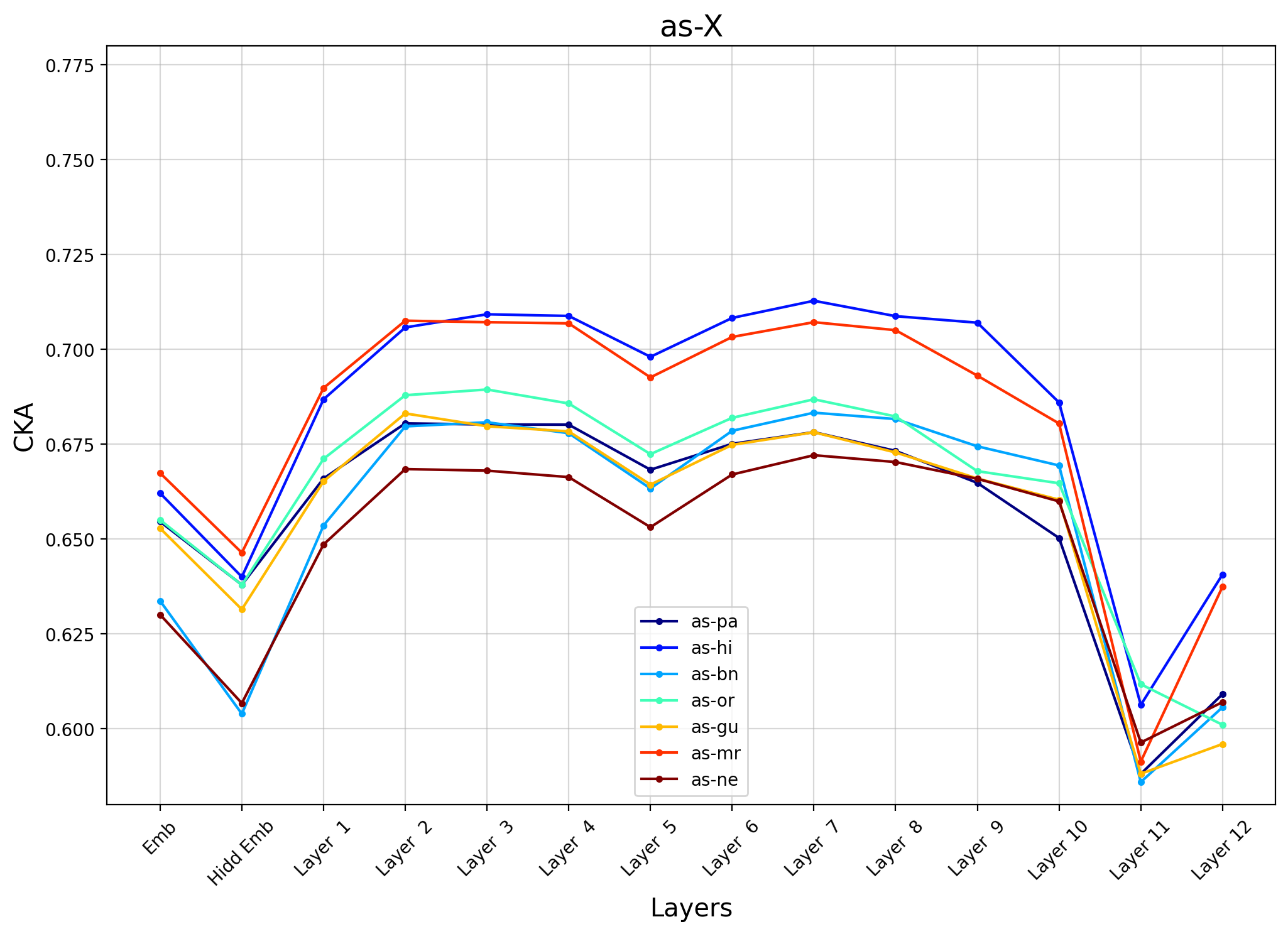}\label{fig:ckaperlangbaseline_as}}
\hfill
\subfloat[uni-script AS-X]{\includegraphics[width=0.40\textwidth]{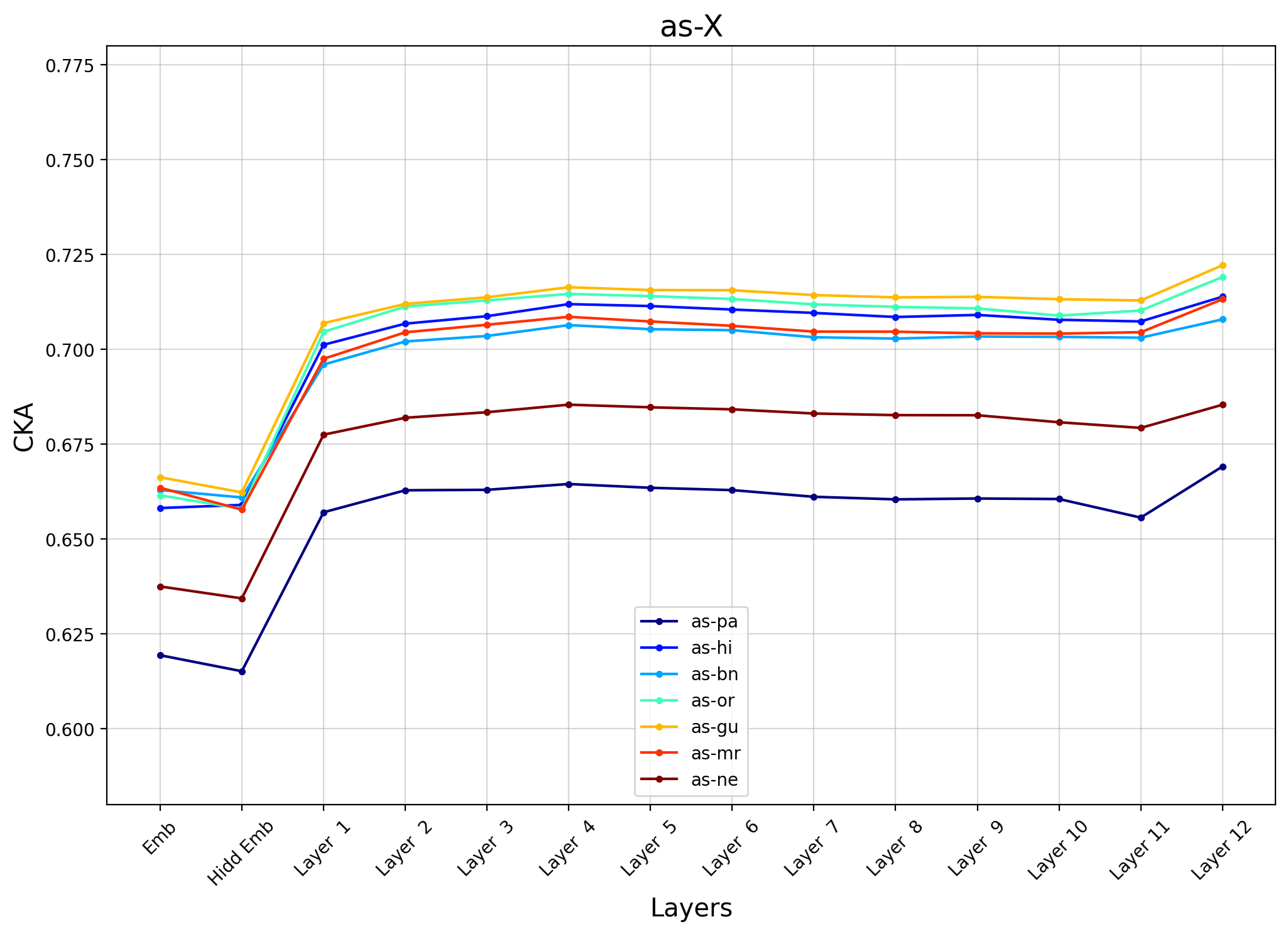}\label{fig:ckaperlangxlmindic_as}}
\hfill
\subfloat[multi-script GU-X]{\includegraphics[width=0.40\textwidth]{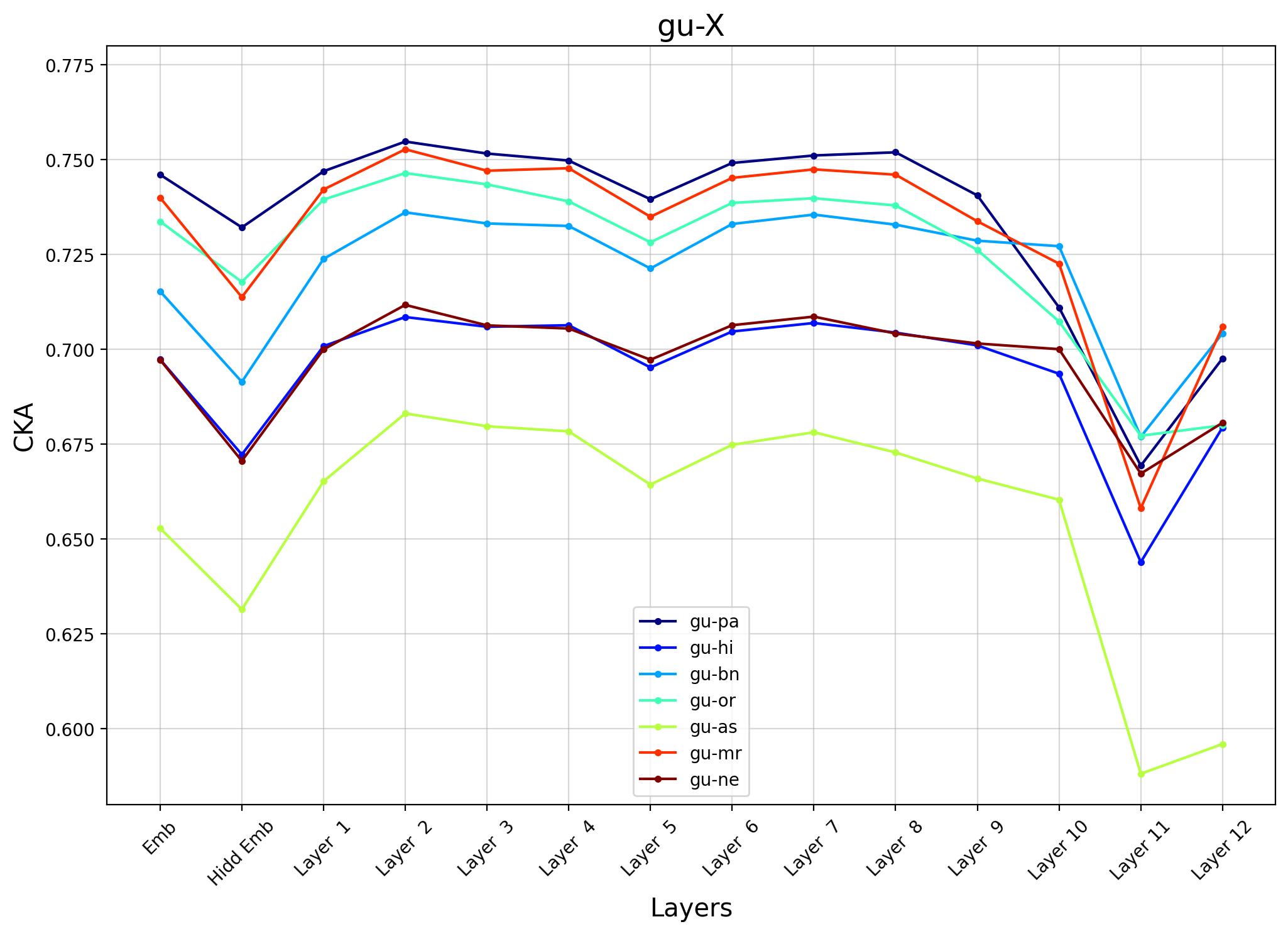}\label{fig:ckaperlangbaseline_gu}}
\hfill
\subfloat[uni-script GU-X]{\includegraphics[width=0.40\textwidth]{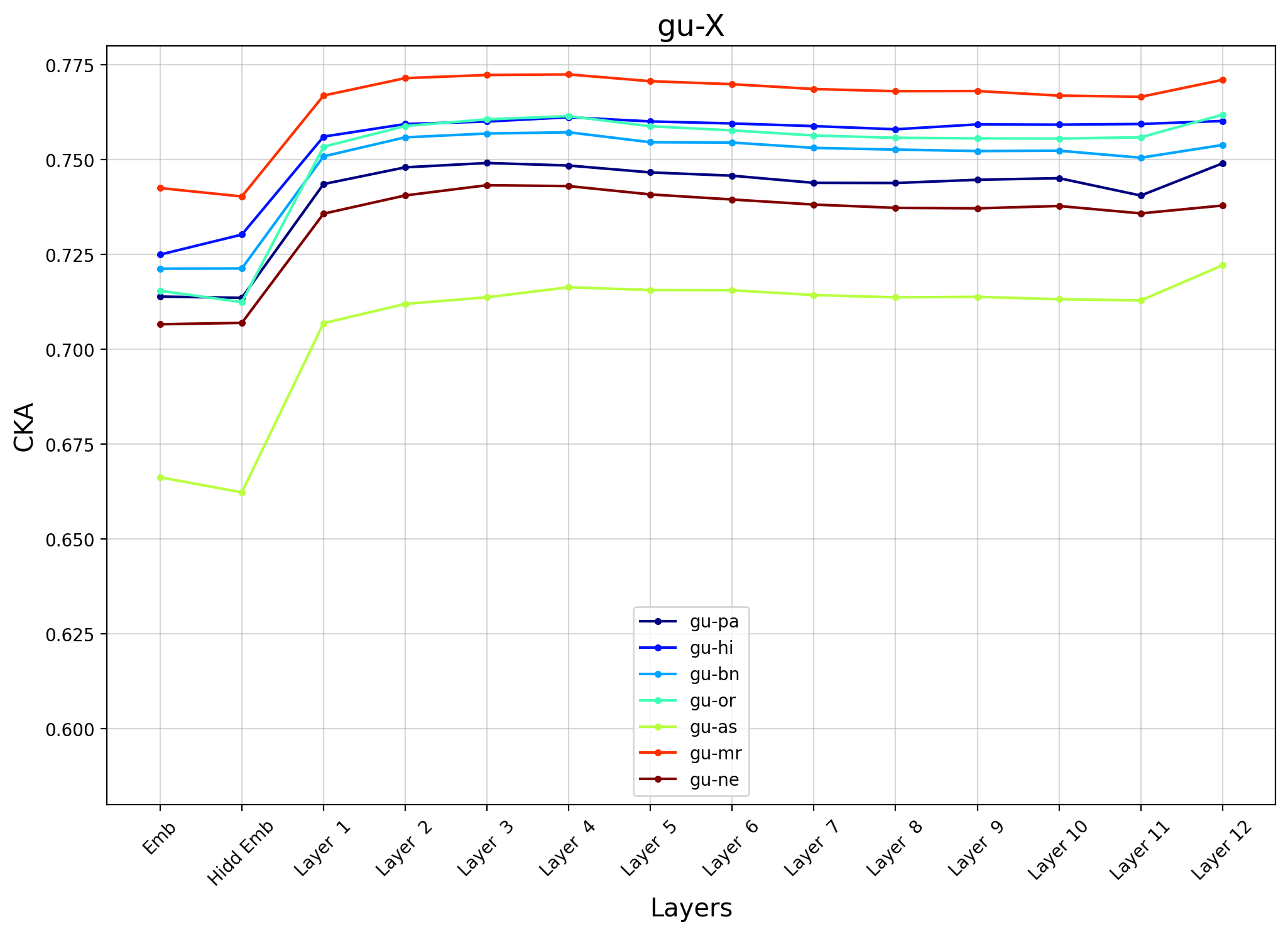}\label{fig:ckaperlangxlmindic_gu}}
\hfill
\subfloat[multi-script MR-X]{\includegraphics[width=0.40\textwidth]{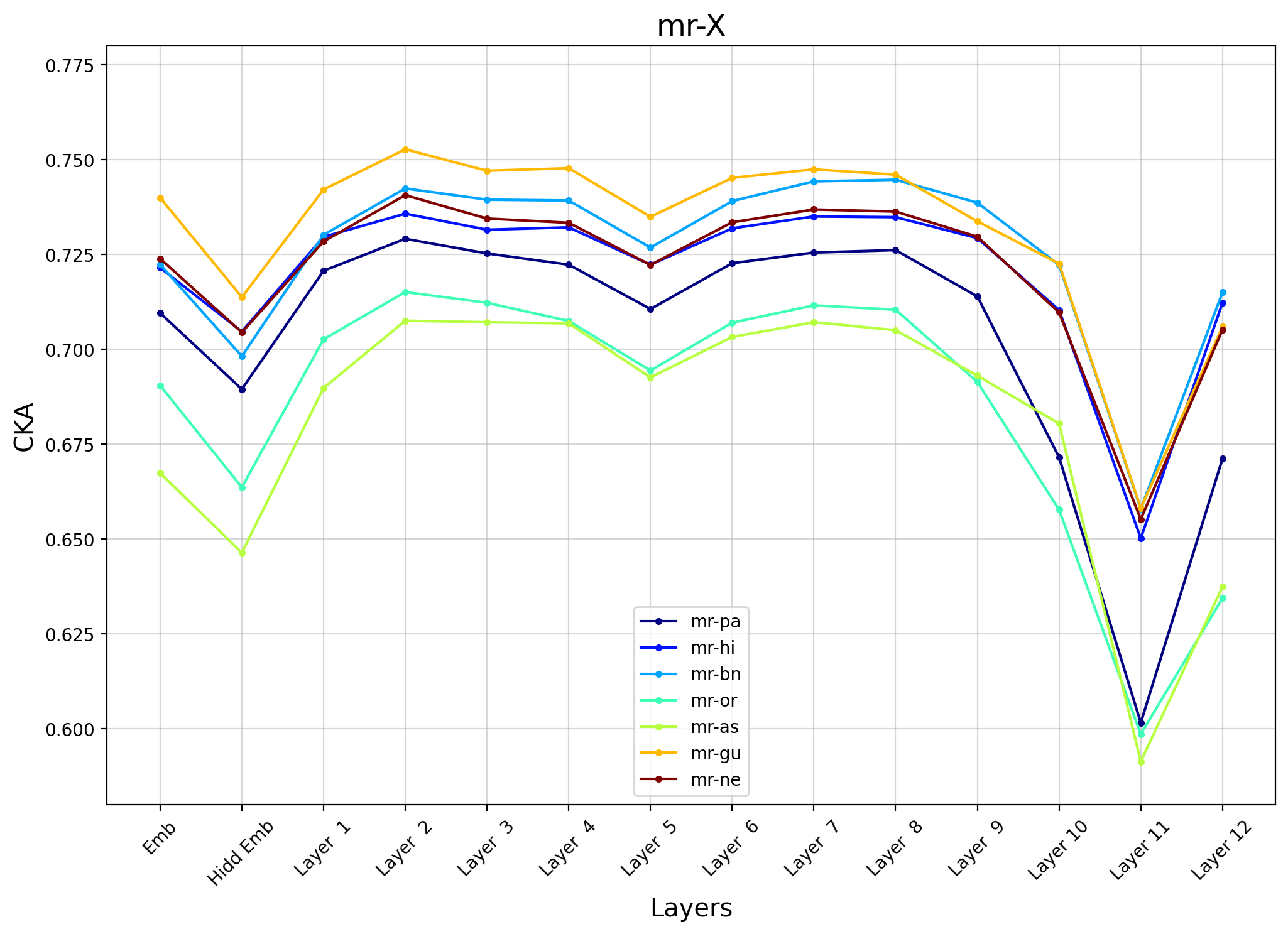}\label{fig:ckaperlangbaseline_mr}}
\hfill
\subfloat[uni-script MR-X]{\includegraphics[width=0.40\textwidth]{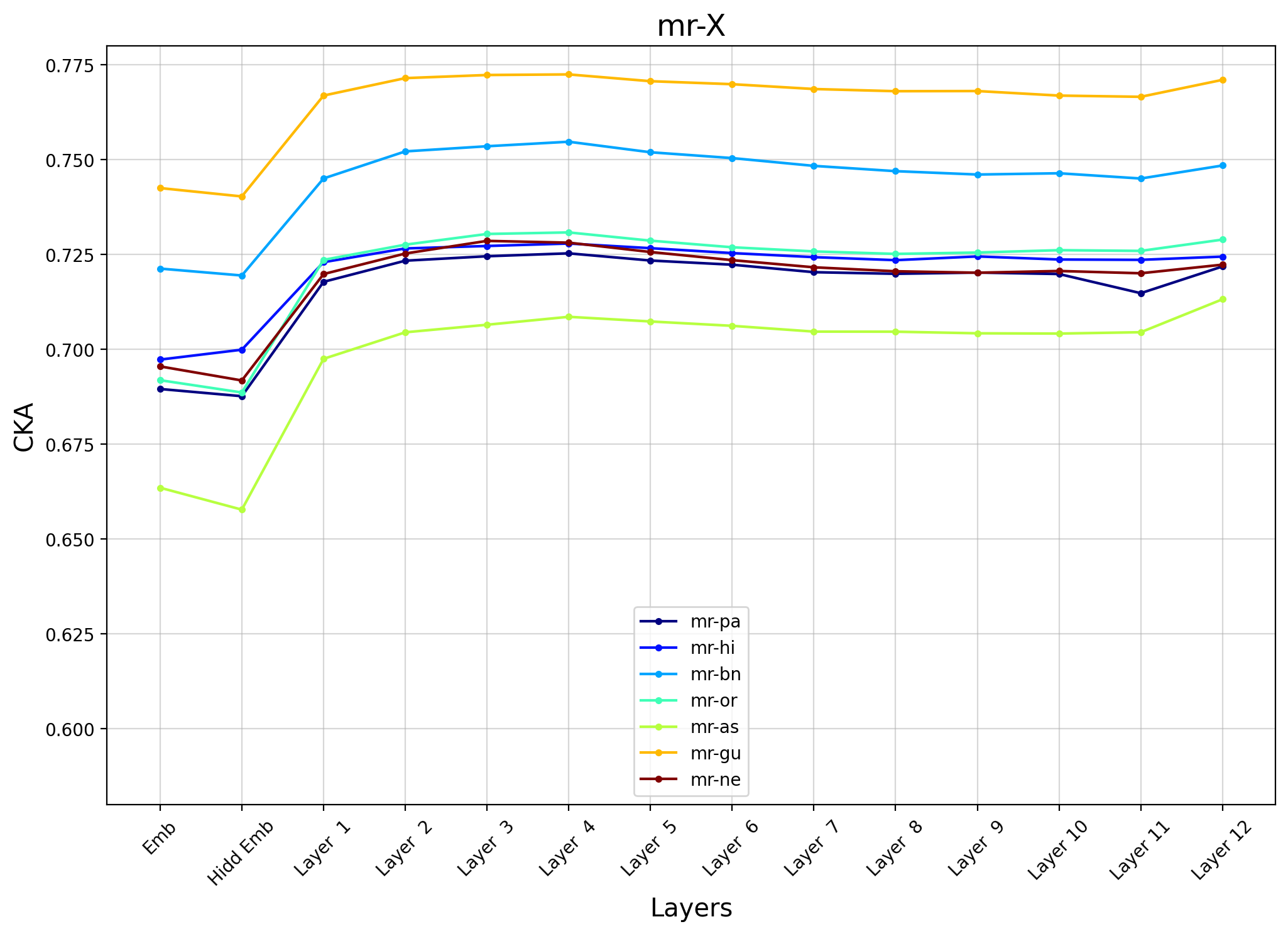}\label{fig:ckaperlangxlmindic_mr}}
\hfill
\subfloat[multi-script NE-X]{\includegraphics[width=0.40\textwidth]{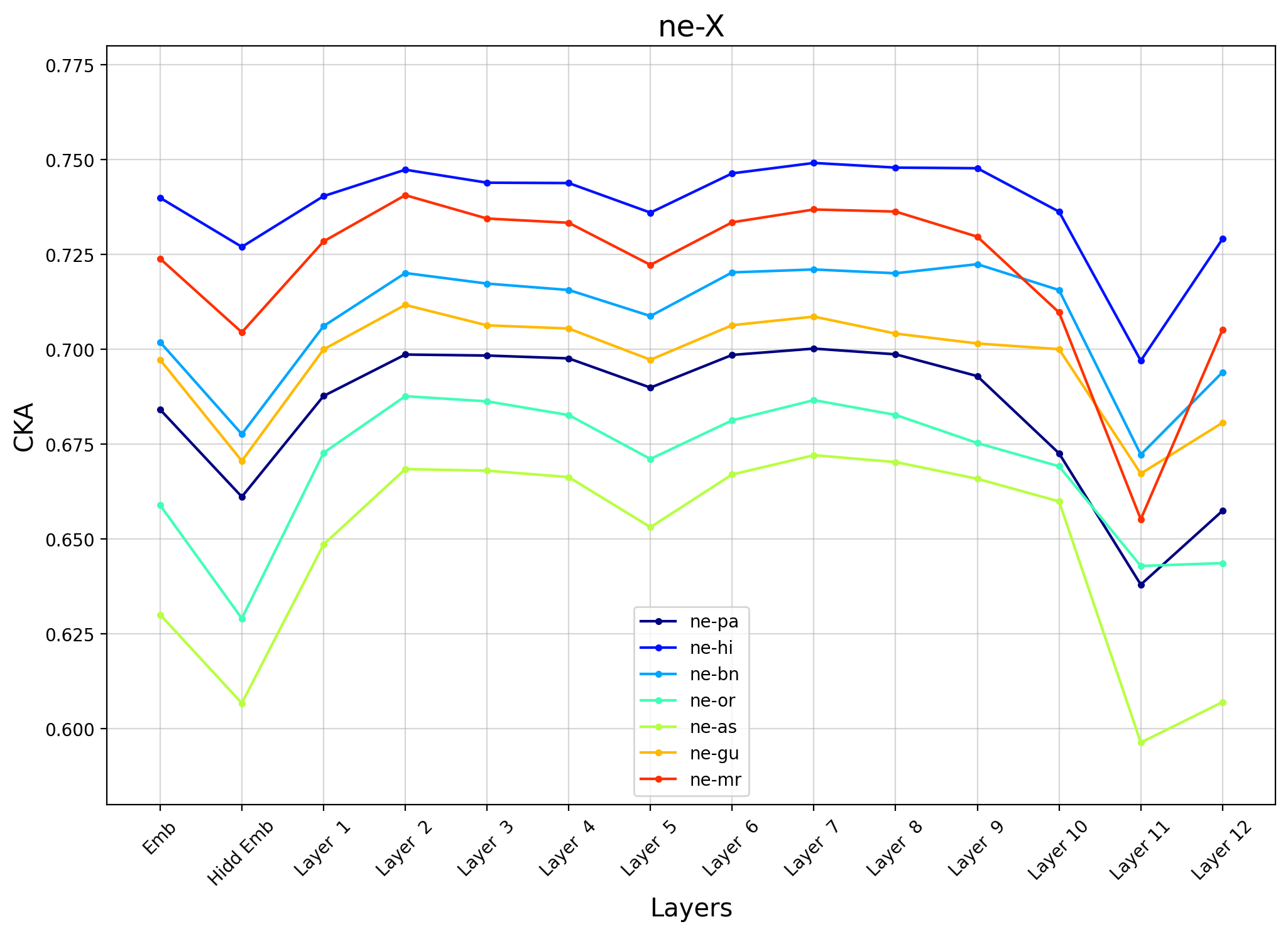}\label{fig:ckaperlangbaseline_ne}}
\hfill
\subfloat[uni-script NE-X]{\includegraphics[width=0.40\textwidth]{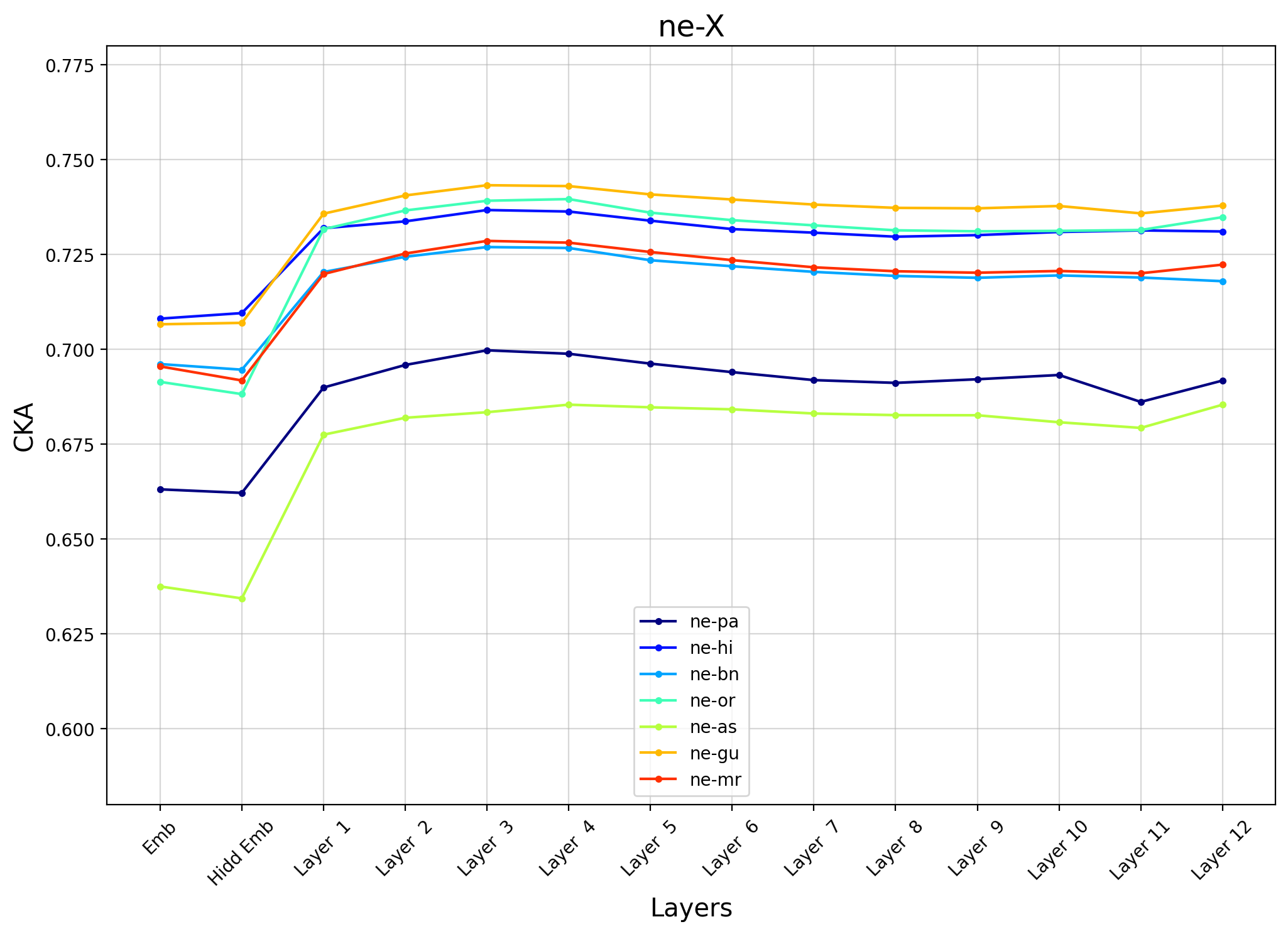}\label{fig:ckaperlangxlmindic_ne}}
\hfill
\caption{CKA of multi-script and uni-script on all language pairs for AS, GU, MR and NE}
\label{cka_all_language_pairs_as_to_ne}
\end{center}
\end{figure*}

\end{document}